# A Novel Black Box Process Quality Optimization Approach based on Hit Rate

*Yang Yang[a], Jian Wu[b], Xiangman Song[c]*, Derun Wu[d], Lijie Su[e] and Lixin Tang[a]*

[a]Key Laboratory of Data Analytics and Optimization for Smart Industry (Northeastern University), Ministry of Education, China (e-mail: yang_cmu@icloud.com; lixintang@ise.neu.edu.cn)

[b]Liaoning Engineering Laboratory of Data Analytics and Optimization for Smart Industry, Shenyang, 110819, China

[c]Liaoning Key Laboratory of Manufacturing System and Logistics Optimization, Shenyang, 110819, China

[d]Manufacturing Department of Shanghai Meishan Iron & Steel Co., LTD., Nanjing, Jiangsu 210039, China

[e]Frontier Science Center for Industrial Intelligence and Systems Optimization

**ABSTRACT**

Hit rate is a key performance metric in predicting process product quality in integrated industrial processes. It represents the percentage of products accepted by downstream processes within a controlled range of quality. However, optimizing hit rate is a non-convex and challenging problem. To address this issue, we propose a data-driven quasi-convex approach that combines factorial hidden Markov models, multitask elastic net, and quasi-convex optimization. Our approach converts the original non-convex problem into a set of convex feasible problems, achieving an optimal hit rate. We verify the convex optimization property and quasi-convex frontier through Monte Carlo simulations and real-world experiments in steel production. Results demonstrate that our approach outperforms classical models, improving hit rates by at least 41.11% and 31.01% on two real datasets. Furthermore, the quasi-convex frontier provides a reference explanation and visualization for the deterioration of solutions obtained by conventional models.

**Keywords**

Data analytics; Hit rate optimization; Process product quality; Quasi-convex optimization; Bi-level optimization; Hidden Markov process; Multitask sparse learning;

**ABBREVIATIONS**

AE, autoencoder; BF, blast furnace; BOF, basic oxygen furnace; MTEN, multitask elastic net; HMM, hidden Markov model; HMT, hot metal temperature; HRO, hit rate optimization; LASSO, least absolute shrinkage and selection operator; MTLASSO, multitask least absolute shrinkage and selection operator; OLS, ordinary least squares; SVM, support vector machine.

## 1. INTRODUCTION

Data analytics under Industry 4.0 has significantly facilitated the smart factory or the integration of industrial processes [1]. The interaction between the real and the virtual is being revolutionized, especially the control of physical entities in reality with the virtual. For the perceptual and predictive performance of production system, the validity and applicability of dynamic simulation models are particularly important. At this stage of industrial evolution, the model parameters in the data analytics production system is becoming more controllable and reliable, leading to higher levels of automation, integration and functional modularity; synchronously, taking into account the superior interaction, these technological advances will guide technologists to be further aware of the safety boundaries of the automated production, the mechanism potential of the integrated production and the optimization of the coordination between functional modules.

Integrated steel production is a typical example of the industrial process integration [2]. And, autonomous steelmaking is also a prime instance of the data analytics production system. Typically, there is a prescribed range of product quality from the upstream process output to the downstream process, referred to as the hit range of the process product quality. Meanwhile, the downstream process holds a carrying capacity for receiving the product quality outside the hit range, which can be evaluated by a metric termed hit rate. In addition, the multi-departmental manufacturing need to consider the optimization of hit rate, energy, and other whole-system indicators while predicting and controlling process product quality. The integrated steelmaking requires the participation of several production departments to coordinate and optimize a dynamic cyber-physical system for the blast furnace-basic oxygen furnace (BF-BOF) process [3]. As shown in Fig.1(*a*), the molten iron dispatch is the central tie of the BF-BOF process route, connecting and coordinating the upstream and downstream processes. The molten iron is injected from the BF of the ironworks into a torpedo ladle. Then, it is transported by a railway locomotive to the BOF of the steelworks. Finally, the empty ladle is sent back to the BF to complete the BF-BOF route. The BF-BOF route involves complex physical and chemical processes, including the gradual wear of refractories and various environmental conditions [4]. In addition, the continuous development of steel recycling is bringing new challenges to the molten iron dispatch [5]. The addition of recycled steel leaves the entire process exposed to the unknown risks in terms of production safety and energy control. Energy costs and operational difficulties are hanging in the balance to ensure hit requirements.

In order to coordinate the above-mentioned multi-departmental production operations and optimize the process-wide indicators, the data analytics system needs to meet stringent hit rate requirement for perceiving and predicting the process product quality. This article provides a novel research perspective on convex optimization-based data-driven approach for the autonomous integrated steelmaking in Industry 4.0. As shown in Fig.1(*b*), the data analytics techniques used in hot metal scheduling consist of two categories: the dynamic regularized surrogate modeling for process temperature simulation [See Sec. (3.1)]; and the data analytics models based on virtual stochastic processes and statistical inference when making high frequency new production decisions to interact and control steel production [See Sec. (3.2)]. In this case, each torpedo ladle is recognized separately as a data carrier, and the state of the process product is hidden. The consequential dataset problems pose practical challenges to the data-driven modeling, such as dataset shift [6], small sample dataset [7], and dataset imbalance [8]. Coupled with limited or incomplete information about intermediate processes, the direct employment of mechanism or artificial intelligence models is not sufficient to provide a viable solution. As a result, these challenges become an obstacle to building and optimizing a completely autonomous smart factory today. To break through this obstacle, the process product quality prediction based on the hit rate indicator has been recommended as the cornerstone for data analyticsing and optimization of the BF-BOF process route.

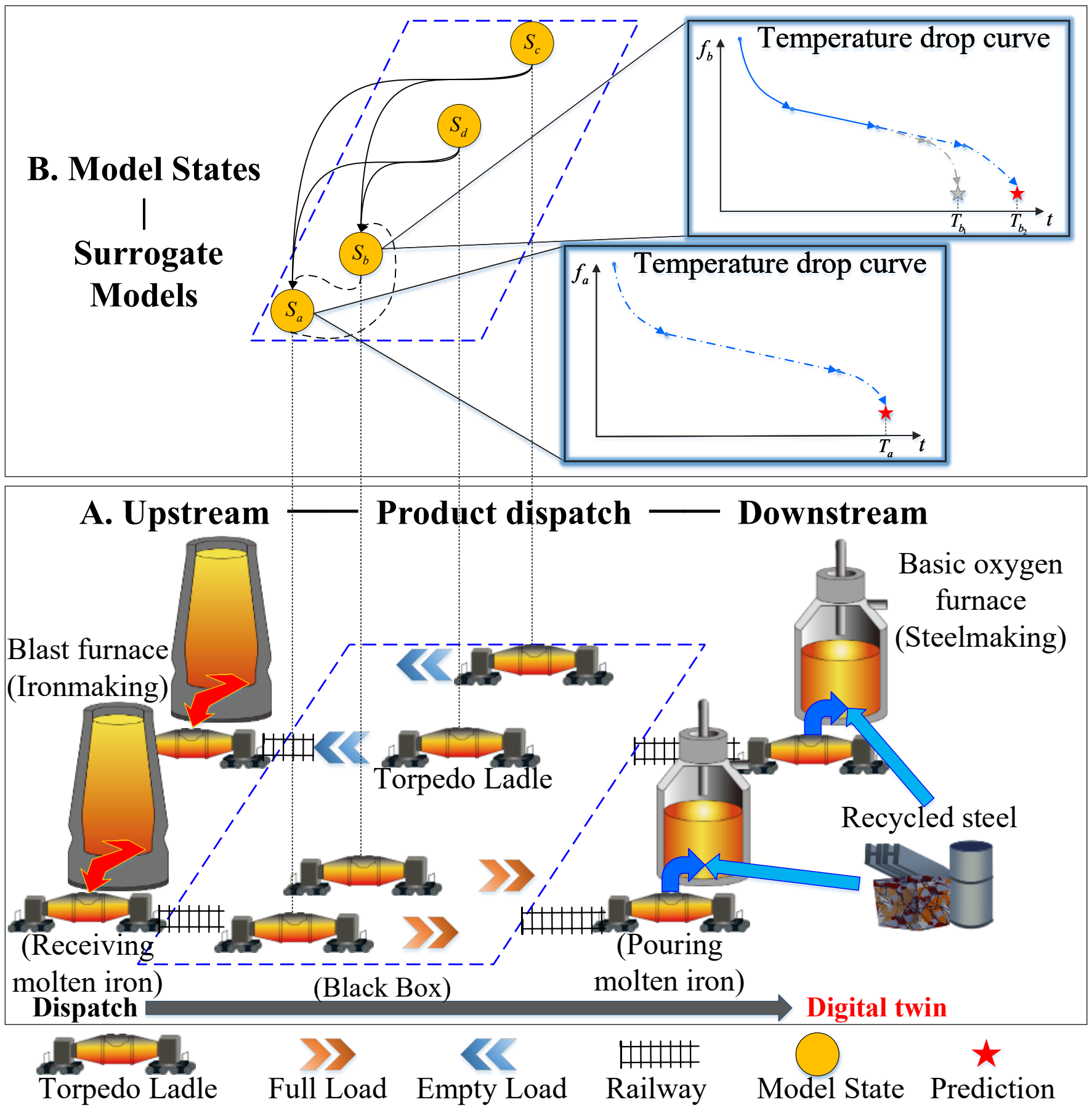

**Fig. 1.** Molten iron dispatching of autonomous steelmaking in Industry 4.0: *a*) the schematic diagram of the BF-BOF route; *b*) the data analytics of the hot metal temperature.

Hot metal temperature (HMT) is an important dimension of the process product quality in the integrated steel production. The safety, efficiency and stability of routine operations mainly depend on the simulation of the HMT curve at the time of molten iron dispatching. Considering the process product quality modeling in integrated steelmaking, the next generation of data analytics technologies should ensure the validity of the simulation models under the feasible hit rate requirement for different integrated industrial processes and explore the optimal hit rate, rather than optimizing individual processes. Therefore, the hit rate optimization of the HMT prediction is not only applicable to the actual requirements of steel production, but also can provide more accurate and flexible coordination of the scheduling time and process parameters for the BF-BOF route [9]. In the dynamic cyber-physical system of the data analytics, the hit rate optimization emphasizes the validity and practicability of simulation results for production decisions and process control [10]. In contrast, although single product quality model optimization or single process optimization has the optimal performance in theory, it is difficult to implement or validate in actual whole-system production due to restricted conditions and limited costs. As a result, the dynamic optimized hit rate indicator with real-time data analytics will serve as a key criterion for the new generation of the data analytics technologies.

Summarized above, the process product quality modeling and optimization are regarded as two critical research topics for the data analytics of the integrated process industry.

The mechanism-driven modeling and data-driven modeling are the dominant approaches for industrial data analytics to perceive the knowledge on the process product quality. As a first step, the mechanism research is adept at exploring and providing the most direct scientific explanation on the relationship between the process product quality and the control variables of the industrial production. The complex physicochemical processes involved in the molten iron dispatch have been studied by mechanism-driven simulation models [11-13]. The mechanism-driven simulation displays the specific macroscopic variations of each product, taking into account the differences in their microscopic properties. Nevertheless, the mechanism-driven simulation depends on high-performance computational resources and enormous training time, which is not suitable for real-time applications.

In order to complement the mechanism-driven models, the data-driven simulation is identified as a promising orientation [14,15]. Artificial intelligence, big data analytics with optimization [16,17], and statistical learning and inference [18,19] are all recent popular data-driven modeling approaches. With regard to black-box modeling with the introduction of physical constraints and control conditions, the theoretical research and industrial applications of artificial intelligence methods are currently in the exploratory stage [20,21]. Aiming at the black-box modeling problem, mixed integer programming models and threshold shrinkage methods perform well in high-dimensional data analytics. are two promising optimization approaches for high-dimensional process data. A software for automatic learning of algebraic models (ALAMO) is developed by a team led by professor Sahinidis, which relies on mixed integer linear programming (MILP) techniques to implement high-precision and low-complexity surrogate models [22]. Network functional varying coefficient model provides the frontier data analytic with statistical analysis, though it requires statistical a priori knowledge of the coefficients [23]. Based on the stronger generalization capability of functional regression, the optimal combinations of smooth basis functions are investigated for actual industrial processes [24,25]. Compared to the explicit models, they need more exploration in data analytics. Based on the functional regression, the optimal combinations of smooth basis functions have been explored for describing complex high-dimensional process industrial systems, and efficient threshold shrinkage algorithms for parameter estimation have been proposed by a team led by professor Shi [24]. Data-driven methods usually require sufficient data. However, this condition is difficult to achieve in the process industry because the same equipment often produces many different grades of products. In addition, there is often a contradiction between actual field requirements and model prerequisites. Higher model accuracy always means higher production costs, higher operating standards and more reasonable mathematical assumptions. These are more challenging to achieve for the further upstream processes in the process industry. Therefore, the main task in today's industrial renovation is to find appropriate compromises and to perceive the production potential of existing processes based on the comprehensive process hypothesis that balances safety, capacity and hardware conditions. In this regard, hidden Markov models (HMMs) and regularized multitask learning methods are two inclusive options for modelling the stochastic process on actual datasets. The HMMs extensively analyze prior relationships among process data and employ the relationships to establish variational models [26,27]. The multitask learning is used to improve the accuracy of the models by sharing useful information among multiple relevant grades and tasks [28]. However, the process industry problems are fundamental and increasing uncertain, and models have to inevitably rely on simplifying assumptions, inaccurate predictions as well as limited or incomplete information [29]. Therefore, the purposeful optimization considering data characteristics and project scenarios is essential for the data-driven modeling approaches.

In terms of the optimization, both optimization-based modeling and model-based optimization are common optimization procedures. The model-based optimization includes black-box optimization, robust optimization, probability optimization with confidence interval, ensemble optimization methods and etc. Typically, the black-box optimization approach is dependent on input-output relationships based on a large amount of data, which values the prediction accuracy of the terminal product quality and serves the process control [30]. The robust optimization utilizes the expected risk minimization as a process-wide optimization strategy [14,15]. The probability optimization evaluates and improves the fluctuation range of the dynamic product quality from the perspective of probability [31,32]. The ensemble optimization lies in assembling multiple shallow models. When it comes to the optimization-based modeling, the optimization procedures highlight optimization techniques for the absence of an explicit model representation [33,34].

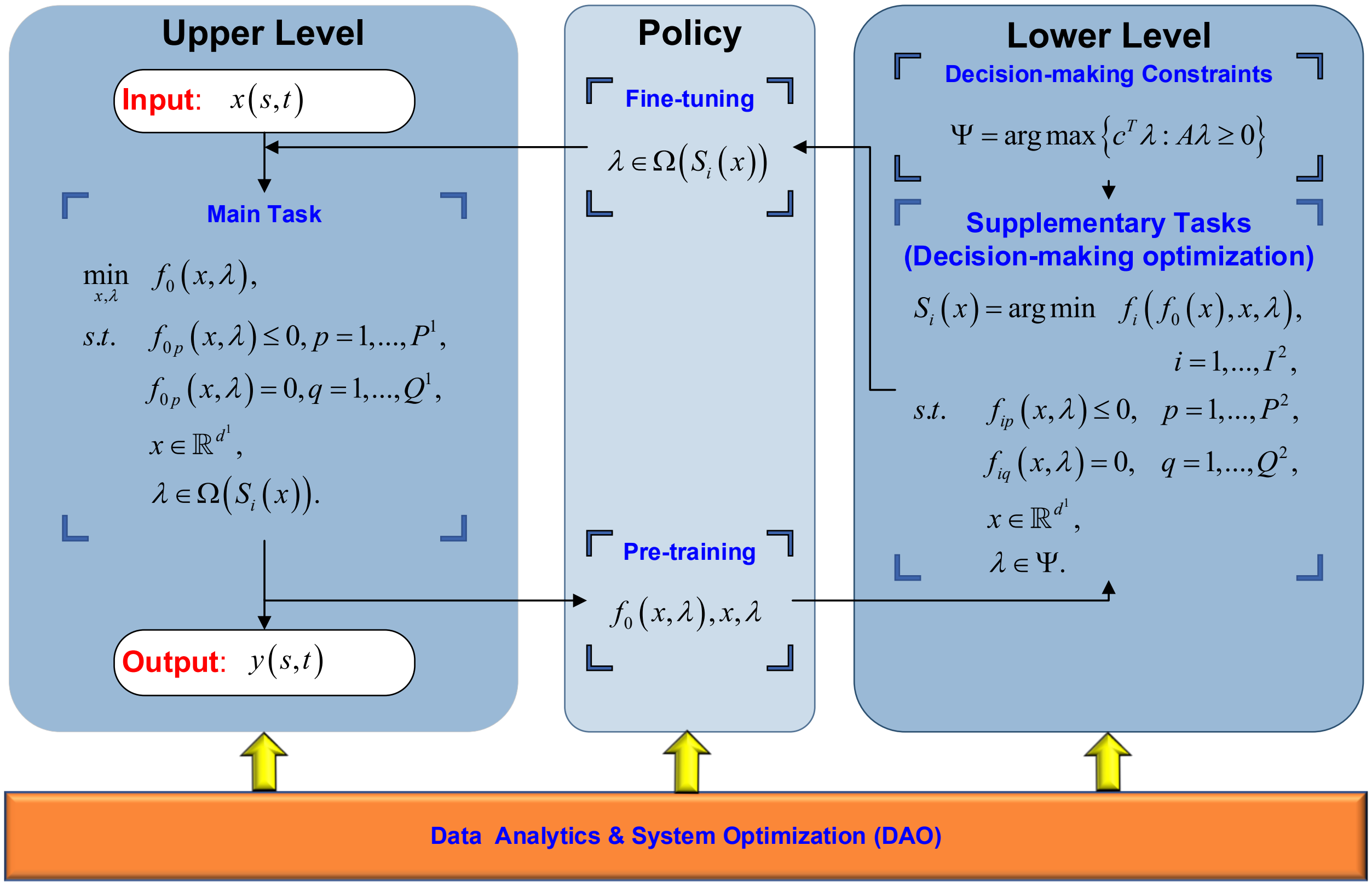

**Fig. 2.** Data analytics and optimization framework based on bi-level optimization.

Fig. 2 presents a bi-level optimization framework that employs a quasi-convex optimization method to iteratively optimize upper and lower-level problems of the quasi-convex function with coupling variables. In the context of industrial process integration, it is essential to construct and dynamically optimize data-driven predictive control models online to balance the performance between product quality control and data analytics models [35]. The hit rate optimization method proposed in the DAO framework aims to strike a trade-off between the complex industrial requirements and constraints (lower-level problem) and algebraic statistical models based on sparse optimization (upper-level problem).

Motivated by the successful data-driven surrogate modeling approaches and convex optimization application in previous research, in this paper we investigate an evaluation function based on the hit rate and hit interval for perceiving, predicting and controlling the process product quality in data analytics production systems. However, different from the previous mixed modeling and optimization approaches, the proposed data-driven quasi-convex optimization approach has the following features.

1) With respect to the integration of industrial processes, the process-wide optimization requires the exploration into controllable process product quality modeling and prediction methods for the flexible coordination between upstream and downstream industries. The hit rate optimization problem of process product quality and the related surrogate models are investigated based on the hit rate indicator. The proposed data-driven modeling and optimization approach is utilized to optimize or control the predicted upstream product quality and simultaneously generates an explicit functional expression of the product quality. The proposed approach has been successfully applied in the integrated steelmaking and offered optimal or controllable hit rate with corresponding operating schemes. The numerical results demonstrate that the proposed approach contributes to the stability, safety, accuracy, and interpretability of the data analytics production system.

2) Further exploring the above industrial problem, the efficiency of dynamic modelling and dataset shift problems are revealed. The regularized surrogate model and the quasi-convex optimization algorithm of the controllable hit rate supply the options for the technical experts to decide the quality with the efficiency, whether to pursue the highest quality or to balance the optimal conditions for both efficiency and product quality.

3) In terms of mathematical optimization, the optimization for the hit rate of product quality models is a sophisticated nonconvex intractable problem. On the one hand, based on the surrogate convex approximation modeling, the nonconvex problem is transformed into a set of convex feasible problems; on the other hand, the quasi-convex optimization algorithm is proposed to optimize the convex feasible problems. Consequently, based on the convex programming approach, not only the optimal hit rate solution can be guaranteed but also the hit rate can be controlled in the feasible domain and suitable product quality models can be inferred.

4) Further exploring the hit rate optimization problem, during the simulation experiment of the quasi-convex optimization, we found that the profile data based on the hit rate controlling process contains a quasi-convex front with respect to the multi-objective optimization of the process variables. The profile is named as the quasi-convex frontier. This frontier has a definite physical meaning, which not only enhances the data interpretability for mathematical optimization, but also improves the natural perception of the data analytics production system.

5) Last but not least, the proposed hit rate optimization approach is a data-driven modeling and optimization framework that allows most of the traditional performance indicators to be integrated into it.

The remainder of this article is organized as follows. Section II introduces the proposed hit rate optimization problem and the concept of dataset shift for the black-box process in process industry. In order to address the hit rate optimization problem, the dynamic regularized surrogate modeling approach and the quasi-convex optimization approach are presented in Section III. Section IV elaborately designs and introduces several simulation experiments and real industrial cases, respectively. The computational results of the experiments are reported and analyzed. The advantages of the HRO approach are demonstrated by comparing it with other methods. Coupling the results of the two kinds of experiments, Section V discusses the essential reasons for the hit rate optimization to improve the process industry based on the schematic diagram of full process optimization. Finally, Section VI presents some conclusions and prospects.

## 2. PROBLEM FORMULATOIN

### *2.1 Upper-Level Problem: Modeling with Hit Rate Optimization*

The hit rate indicator provides the critical assessment for both qualitative and quantitative analysis in the process industry. For the former, the hit rate is usually treated as a static constant metric $\theta^{[h]}$. The research stage of this article is to move from the static hit rate to a dynamic model and then to guide the real-world operations based on the optimized hit rate model (*refer to Li* [1], *p*. 4, *Fig*. 1). The static hit rate indicator is defined as:

$$\theta^{[h]} = \frac{n_q}{n_p} \times 100\% , \tag{1}$$

$$n_q = \sum_{i=1}^{n_p} \mathbf{I}\left(y_i^q\right), \tag{2}$$

$$\mathbf{I}\left(y\right) = \begin{cases} 1, & b_l \le y - y_i^* \le b_u ; \\ 0, & otherwise. \end{cases} \tag{3}$$

where $n_q$ is the number of the qualified products and $n_p$ is the total number of products. $y_i^q$ is the qualified standard value of product quality and $y_i^*$ is the actual product quality output in the final stage. $\mathbf{I}(x)$ is an indicator function to calculate whether the deviation is within the required hit range. $b_l$ and $b_u$ represent the lower and upper bounds of the hit range given by the downstream process.

As the crucial standard for the data analytics production system, the further development of the static hit rate indicator is the dynamic hit rate modeling and optimization. The dynamic hit rate model is a non-convex optimization function based on the relationship between the independent variables and the indicator function, which is designed to assess and control the uncertainty associated with high frequency production decisions and to assist in real-time product scheduling throughout the entire supply chain. In addition, the hit rate works both before/after the production paths (process control variables) are determined and the final product quality is formed. In order to dynamically optimize and simulate the hit rate and process product quality, we propose an optimization-based modeling approach that consists of two alternating phases: optimization and modeling, as shown below.

*Phase I*:

$$\theta_t = \max_{\beta} \mathbf{H}(\beta; x_i), \tag{4}$$

$$\text{or } \beta = \arg\min \left\| \mathbf{H}(\beta; x_i) - \hat{\theta}_0 \right\|_1, \tag{5}$$

$$\mathbf{H}(\beta; x_i) = \frac{\sum_{i=1}^{n} \mathbf{I}\left(f_0(\beta; x_i) - y_i\right)}{n} \times 100\%, \tag{6}$$

$$\begin{aligned} s.t.\ & \theta_l \le \mathbf{H}(\beta; x_i) \le \theta_u, \\ & \exists\, \theta_l,\, \theta_u \in \mathbb{R}, \\ & 0 \le \theta_l,\, \theta_u \le 1. \end{aligned} \quad , \tag{7}$$

The hit rate optimization problems correspond to two paths: *a*) the process data analytics problem of exploring the upper limit of the hit rate; *b*) the operational optimization problem of controlling the hit rate curve to optimize the process variables. The related hit rate optimization problems are respectively formulated in Eq. (4) and Eq. (5), and the hit rate model is defined in Eq.(6). $\theta_t$ represents the average hit rate in one production plan/batch, and $\beta$ represents the real-time optimized parameters of the surrogate model (quasi-convex function) $f_0(\beta; x_i)$ for the process product quality function in the hit rate model. $\hat{\theta}_0$ is the hit rate requirement given by the downstream process. $x_i$ represents the process variables, which includes both the control variables $u(t)$ and system state $s(t)$, and is discretized over time. $y_i$ represents process product quality of the $i^{th}$ product, respectively. $n$ is the total number of manufacturing products in the production plan/batch.

In Eq. (7), the upper bound $\theta_u$ and lower bound $\theta_l$ of the hit interval usually need to be optimized and predicted as well. The hit interval is also the value range of $\hat{\theta}_0$. Therefore, the lower bound of the hit optimization process can be raised based on the empirical estimation $\hat{\theta}_l$ in practice.

Let $f(x_i): \mathbb{C} \to \mathbb{R}$ be a $C^2$ (or twice continuously differentiable) function that represents the original product quality function, where $\mathbb{C}$ is a $C^2$-Riemannian manifold (*refer to Lafontaine* [36], *p.* 10). The next step is to find the surrogate model $f_0(x_i^d): S_d \to \mathbb{R}$, $S_d \subset \mathbb{R}^d$ that is the approximate real-valued algebraic model to the original function $f(x_i)$ on $C^2$, where $x_i^d$ represent a set of $d$ covariate functions for $i^{th}$ product, which also determines the dimension of the optimal parameter $\beta$ and the process variables of the surrogate model $f_0(\beta; x_i)$ in the first phase. $y_i^*$ represents the actual product quality received by the downstream process.

*Phase II*:

$$\textbf{find} \quad u(t) = \arg\min L(u(t), s(t), \beta) \tag{8}$$

$$s.t. \quad f_0(x_i^d) \approx f(x_i) \approx y_i^*, \tag{9}$$

where Eq. (8) presents an optimization problem for the loss function in a predictive control model for process product quality. To optimize the loss function, it is important to ensure that the surrogate function is a convex approximation that closely approximates the true process product quality.

During the second phase, it is important to design a feasible and reliable surrogate model for the process product quality based on the optimization model in the first phase, so as to guarantee the convexity and interpretability of the hit-rate optimization model.

The optimization-based modeling problems in phase I and phase II are two significant components for the data analytics of the process industry. In the first stage, it is assumed that the exact model representation of the product quality is unknown. Therefore, it is concerned with finding the feasible model representation according to the optimization problem of the industrial processes. The feature selection and data-driven model representation for different scenarios are provided in the second phase. Then, returning to the first phase, the model parameters are specifically optimized according to the different input data. For the process industry, it is a matter of finding the adequate data-driven models from the complex cyber-physical network [35]. In the second phase, it is also regarded as an optimization problem of process product quality prediction for improving the hit rate. Generally, for the production schedule with different product levels [37], the optimization problem of the hit rate model is intractable and non-convex. Under Industry 4.0, the hit rate model is required to be real-time, controllable and predictable [38]. By this way, the dynamic data-driven modeling and optimization problems based on the sample efficiency, generalizability, model composition and incremental updating have aroused extensive concern [39].

Considering the intractable but common dataset shift problems in the process industry, this article attempts to propose a unified framework based on a multilevel optimization approach [40], so that the hit rate optimization, the surrogate modeling of the process product quality, the stochastic process modeling of the production process, and the clustering and classification problems of the datasets can all be put into a unified optimization model.

### *2.2 Lower-Level Problem: Hidden Dataset Shift*

In recent years, dataset shift has attracted wide attention as the promotion of industrial data analytics. The dataset shift problem in machine learning was first conceptualized by Storkey in 2009 [41]. Subsequently, Moreno-Torres and co-authors presented a unified framework regarding the classification of the dataset shift [6]. The detection and resolution of the dataset shifting is an inevitable trend in the integrated process industry. The dataset shifting problems include covariate shifting [42], label shifting [43] and concept shifting [44]. It is found that time-window-based approach is an effective and representative solution, but it is still tricky to solve the concept shifting of the black-box process. Notably, all three types are all involved in molten iron dispatch, which means the data analytics in the black-box process involves data pattern recognition and semi-supervised clustering/classification.

The data patterns of industrial high-dimensional problems are considered complex and challenging to visualize and imagine. Typically, the industrial high-dimensional data within any time window is seen as a mishmash of irregular data streams. Therefore, this section provides a low-dimensional example to visually explain the dataset shifting and clustering problems associated with the hit rate optimization problem. Similarly, in Sec. (4.1), the data analytics of the hit rate optimization process can be better demonstrated based on the low-dimensional HMM process. Fig. 3 shows a sample graph of a Gaussian process with multiple population datasets [45]. It is assumed that the process data of each production task via the BF-BOF route consist of the unobserved multiple population datasets. In general, the sample distribution and noise distribution of the data population from different tasks are not exactly the same in different periods and different operating sequences. Let (*X*, *Y*, *T*) be a three-dimensional coordinate system and each moment *t* corresponds to the production task of one product. Then, the products are described by two-dimensional data (*X*, *Y*), where *X* represents the process variables and *Y* is the output product quality. Each product data $\left(X_{ti}, Y_{ti}\right)$ under its true population is assumed to obey the following distribution:

$$Y_{ti} \left| (X_{ti}, S_i = m) \sim N\left( X_{ti}\beta_m, \left(\sigma^{(m)}\right)^2 \right), \right. \tag{10}$$

where *m* is the classification for all the data populations; $S_i = m$ represents the population of product *i* belonging to the $m^{th}$ data population; for the BF-BOF tasks, $X_t$ represents the process variables and variable $Y_t$ represents the terminal HMT of task *t*; $\beta_m$ represents the model parameter of the $m^{th}$ data population, $m=1,2,\ldots,M$. Comparing Fig. 3(*b*) and Fig. 3(*c*), it is found that both the categories of the sample and the distribution of the population, dynamically change over time. However, the practice of regression modeling of the entire dataset is hardly compatible with this issue. In addition, Fig. 3 demonstrates the reason why mechanism analysis believe that the process quality function should be unique for each product, even though this is not feasible in the industrial data analytics.

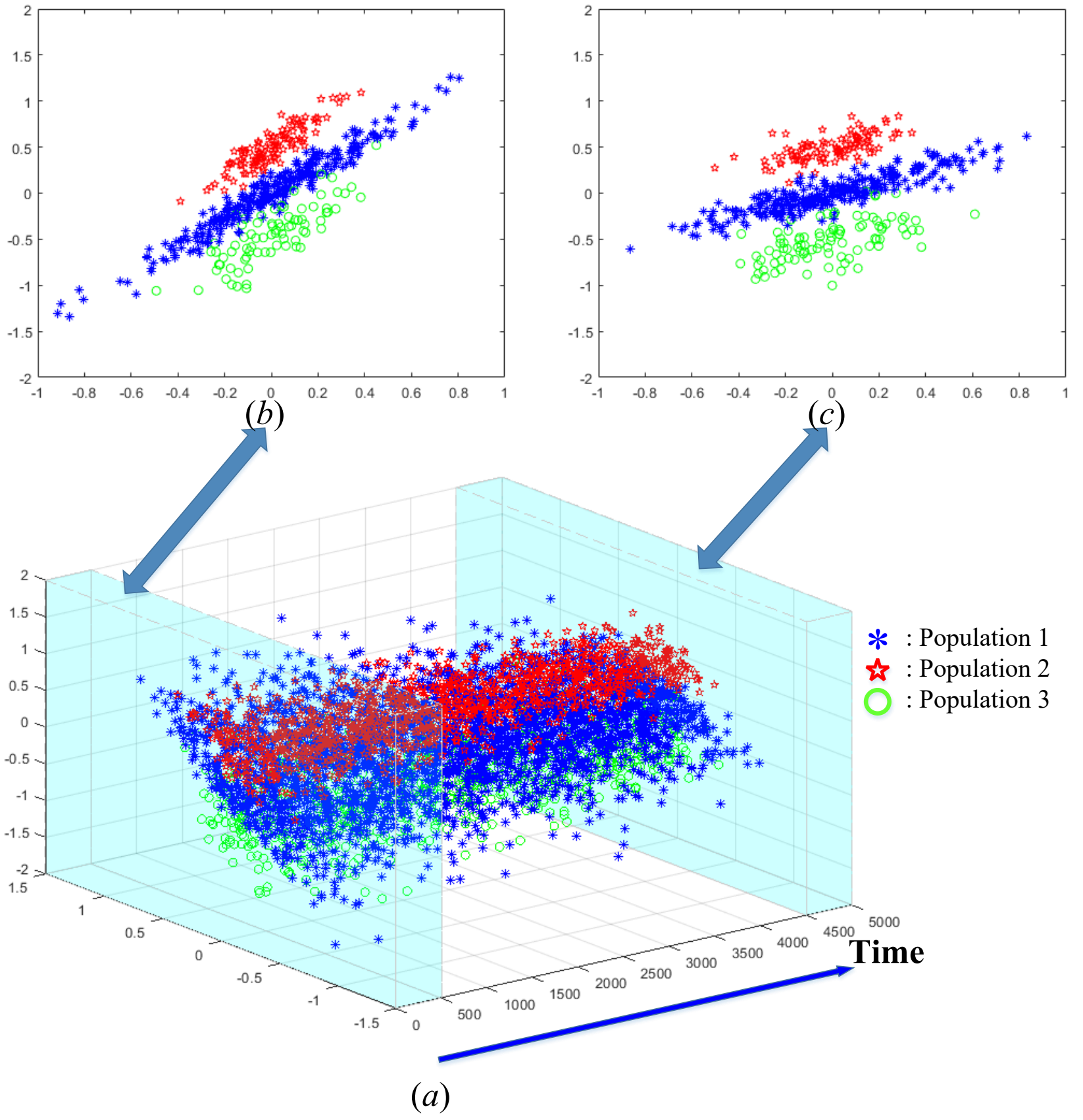

**Fig. 3.** An example diagram of the dataset shift with the Gaussian process; *a*) data samples of 5000 products; *b*) and *c*) represent the projection of the dataset (*X*, *Y*, *T*) during different periods ([0,500] and [4500,5000]) onto the *X*-*Y* plane.

Labeling the black-box data is an inevitable part and challenge in industrial data analytics. The common approaches for black-box data clustering can be divided into supervised, unsupervised, and semi-supervised clustering methods [46]. In the process industry, traditional physicochemical experiments or posterior methods cannot be appropriate to generate the data labels in real time due to the high frequency production schedules and the consecutive production samples between upstream and downstream. In this article, the labeling problem in the hot metal scheduling is discussed from the perspective of mathematical optimization. Based on semi-supervised clustering algorithm and scenario tree method [47], the optimization for clustering is introduced into the multi-level optimization model. Then, the dataset categories are probabilistically divided based a latent set of scenarios $\{S_m\}$ and the optimized parameter $\beta$ of the surrogated models as shown in Fig. 4, and the labeled clusters are dynamically optimized based scenario tree reduction method as shown in Fig. 6 in Sec. (3.2).

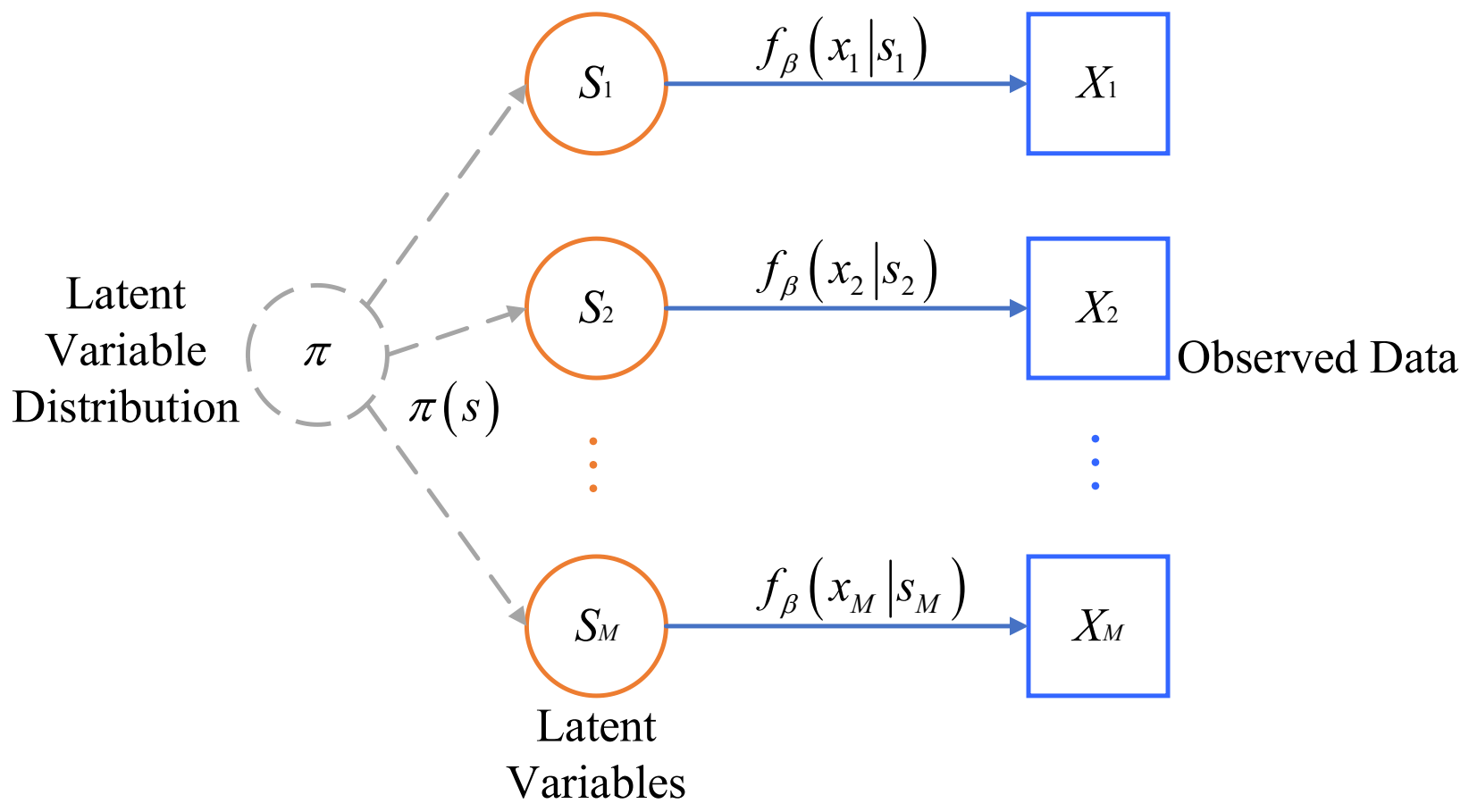

**Fig. 4.** Scenario clustering of latent datasets based on probabilistic models.

Combined with the abovementioned and the authors' existing experience, the hit rate optimization in Eq. (4) and Eq.(5) can be regarded as a kind of special regression problems. In terms of the equivalence between the hit rate and accuracy, for the single-population dataset, if the residuals satisfy the Gaussian distribution and the width of the hit range is small enough, maximizing the accuracy (minimizing the residuals) is considered equivalent to maximizing the hit rate. For multi-population dataset problems, especially black-box processes, there is more research to be done in respect to modeling and optimization of the hit rate model.

## 3. DATA-DRIVEN QUASI-CONVEX OPTIMIZATION METHOD

In this section, the process product quality is discussed under the assumption of the hidden Markov process. The surrogate models are established by multitask sparse learning and their probability are estimated by HMMs. Then, a quasi-convex optimization method is proposed for the hit rate optimization problem. Harmonizing these two models with the proposed quasi-convex framework, a data-driven quasi-convex method is presented for the hit rate optimization, and the optimization process is demonstrated in Fig. 6. The proposed method is designed for the hit rate optimization (HRO) of process quality prediction; therefore, this method is named HRO.

### *3.1 Dynamic surrogate modeling approach with FHMM and regularized feature selection*

A dynamic surrogate modeling and regularized feature selection approach is proposed by combining the FHMM model and ***sparsity techniques*** in Fig. 5. Specifically, the process product quality of different sample clusters (scenarios) $g_m\left(X_t,\hat{\beta}_m\right)$ is estimated using the regularized multitask learning method, and the probability $p_{(S_t=m)}$ of each scenario is expressed by the FHMM model. Through the time window, the dynamic surrogate modeling approach adaptively updates the parameters and optimizes/controls the hit rate curve for different production schedules.

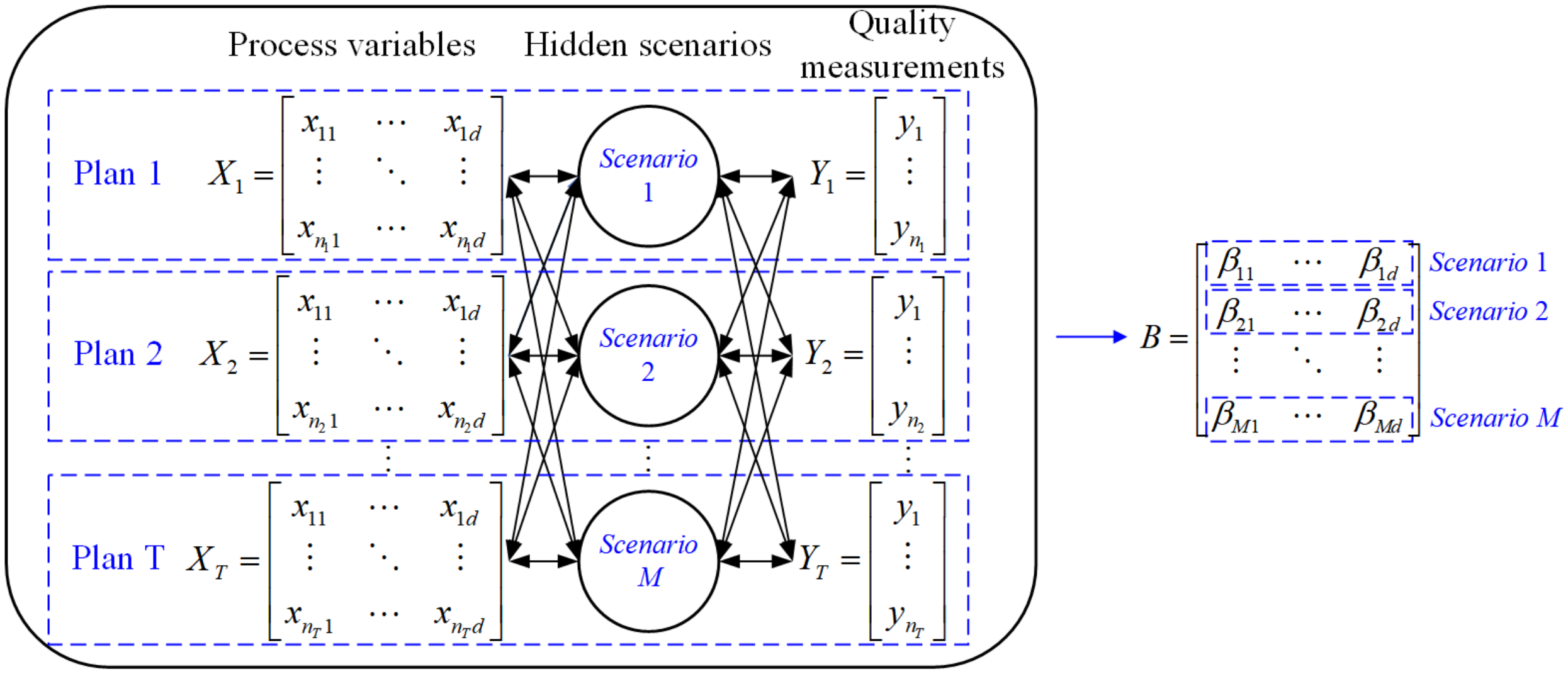

**Fig. 5.** Data-driven dynamic multitask surrogate modeling process based on multiple hidden Markov scenarios.

#### a) Hidden Markov Process and Factorial Hidden Markov Models

As shown in Fig. 1(*b*) and Fig. 5, the equipment state changes over time in the mechanism analysis, which is the dataset shift in the data analytics systems. Let variables and states of the BF-BOF tasks be divided and updated at time *t*, and there are *N* pieces of equipment (torpedo ladles) to complete *Task t*. The relation among the equipment states is assumed to be the first-order hidden Markov process as:

$$p\left(S_t \in \Phi_A \middle| X_1,\ldots,X_t\right) = p\left(S_t \in \Phi_A \middle| X_t\right) \tag{11}$$

$$p\left(O_t \in \Phi_B \middle| S_1,\ldots,S_t\right) = p\left(O_t \in \Phi_B \middle| S_t\right) \tag{12}$$

$$p\left(Y_t \middle| O_1,\ldots,O_t\right) = p\left(Y_t \middle| O_t\right) \tag{13}$$

where $X_t$ is the input variables of *Task t* and $Y_t$ is the output HMT of *Task t*; $S_t$ is a set of predicted model states of the equipment for *Task t*, namely, $S_t=\left\{S_t^{(m)}\right\}$, *m*=1,2,…,*M*; and $\left\{O_t\right\}$, $t=1,2,\ldots$ is the observed sequence of the model state. The forward $O_t$ (*t*=1,2,…,*T*, and *T* is the current completion time) is the model state observed after the completion of *Task t*, and the backward $O_t$ ($t$=$T$+1,…) is the optimized structure of $S_t$ through the clustering process; $\Phi_A$ and $\Phi_B$ are the state spaces of $S_t$ and $O_t$, respectively. And $S_t,\ O_t = 1,2,...,M$, where $M$ represents the maximum number of predicted/observed model states.

According to the hidden Markov process, the state of the scenario-based model corresponding to the temperature drop process of a torpedo ladle is only related to its previous period. Here, the state of the scenario-based model at the previous period is divided into two categories: the model state of the previous task of the same torpedo ladle and the model state of the previous period of different torpedo ladle s. As shown in Fig. 1(*a*), although the intermediate processes are invisible, the HMT (product quality) states are correlated in adjacent time and space of the torpedo ladles (equipment). Then, the surrogated model of the HMT prediction for the $k^{th}$ clustering process is represented by a mixture of Gaussian models as follows:

$$Y_t^{[k]} = \sum_{m=1}^{M^{[k]}} p_{\left(O_t^{[k]}=m\right)} g_m^{[k]} . \tag{14}$$

The probability density for $n_t \times 1$ observation vector $Y_t$ is

$$p\left(Y_t \middle| S_t\right) = \left|\Gamma\right|^{-1/2} \left(2\pi\right)^{-n_t/2} exp\left\{-\frac{1}{2}\left(Y_t - \mu_t\right)' \Gamma^{-1} \left(Y_t - \mu_t\right)\right\}, \tag{15}$$

where $\mu_t = \sum_{m=1}^{M} W^{(m)} S_t^{(m)}$ and $\Gamma$ is the $n_t \times n_t$ covariance matrix.

As the extended HMM models, FHMMs usually represent the state by a collection of $M$ independent Markov chains with different transfer matrices $M_{tr}^{(m)}$ and initial state distributions $\pi^{(m)}$. As shown in Fig. 5, a factorial HMM is used to reveal the hidden states of the equipment. At time $t$, the initial observed state $O_t$ depends on all the hidden state variables $S_t = \left\{S_t^{(1)},...,S_t^{(M)}\right\}$. Their joint probability of the model can be factored as:

$$p\left(\left\{S_t, O_t\right\}\right) = p\left(S_1\right) p\left(O_1 \middle| S_1\right) \prod_{t=2}^{T} p\left(S_t \middle| S_{t-1}\right) \prod_{m=1}^{M} p\left(S_t^{(m)} \middle| S_{t-1}^{(m)}\right). \tag{16}$$

In terms of optimization, the Baum-Welch algorithm, also known as expectation-maximization algorithm (EM), is used to learn the parameters of the FHMM model [48,49]. The E step maximizes the expectation to update the hidden variable (value or probability) of the sample;

the M step updates the probability of the hidden variable under the condition that the hidden variable is determined. Therefore, the EM algorithm of FHMM is also regarded as convex optimization. The optimization problem in step E is shown as follows.

$$\max_{\eta^{new}} Q\left(\eta^{new} | \eta\right) = E\left\{\log p\left(\{S_t, Y_t\} | \eta^{new}\right) | \eta, \{Y_t\}\right\} \tag{17}$$

where $Q$ is an expected log likelihood function of the parameters $\eta^{new}$, given the current parameter estimate $\eta$ and the observation sequence $\{Y_t\}$. $\eta = \left\{W^{(m)}, \pi^{(m)}, p^{(m)}, C\right\}$, where $\pi(m) = p\left(S_1^m\right)$ and $p^{(m)} = p\left(S_t^{(m)} \middle| S_{t-1}^{(m)}\right)$.

**b) Surrogate Modeling based on Multitask Sparse Learning**

Multitask learning has been widely used to estimate the parameters $\beta_t^{(m)}$ of the hidden scenarios in the process. In general, the modeling process is divided into surrogate models and sparsity techniques [50,28].

***Surrogate models.*** Efficient data-driven surrogate models are important for actual steel production. However, it is undeveloped for current deep networks to consider the control conditions and physical constraints in practical industrial problems [51,21]. One recommended step for surrogate modeling is to use the ALAMO method to initially select the variable space. Then, a trade-off is made between the performance of the data analytics model, the factors in the mechanistic model, and the convexity of the model to establish an optimal convex feasible data analytics model of the $m^{th}$ scenario $g_m(x)$. The following section provides an example of using a quadratic programming model as a convex feasible data analytics model. The surrogate model is represented by

$$g_m(x) = \beta_{m,0} + \sum_{i=1}^{D} \beta_{m,i} x_i + \sum_{i=1}^{D} \sum_{j \ge i}^{D} \beta_{m,ij} x_i x_j. \tag{18}$$

where $\beta_{m,0}$, $\beta_{m,i}$, and $\beta_{m,ij}$ are the constant, linear, and quadratic coefficients, respectively. The dimension of the process variables is *D*, and the dimension of the input variables is $d = \frac{1}{2}\left(D^2 + 3D + 2\right)$.

***Sparsity techniques.*** The elastic net model [52] is introduced, which combines the loss function of the ordinary least square model and the penalty functions of LASSO ($l_1$-penalized regression) and the ridge regression model ($l_0$-penalized regression). Then, considering the parameter correlation among the sample clusters and their sample volume, the multitask elastic net (MTEN) is introduced with the group sparsity and the complexity of the coefficients *t* constrained by the $l_{21}$ norm and the squared Frobenius norm as follows.

$$\min_{B_t} L_{MTEN}(B)=\frac{1}{2\sum n_t}\|Y-XB\|_{Fro}^2+(1-\lambda)\|B\|_{21}+\lambda\|B\|_{Fro}^2 \tag{19}$$

where $\lambda$ is a tuning parameter. When $\lambda=0$, the elastic net becomes LASSO regression. And when $\lambda=1$, it is ridge regression; $\|\cdot\|_1$ and $\|\cdot\|_2$ are the $l_1$ norm and $l_2$ norm, respectively, which coordinate the complexity of the coefficients. $\|B\|_{21}=\sum_{m=1}^{M}\|\beta_m\|_2$ is the group sparse penalty. $n_t$ is the number of the data samples in *Task t* and $\sum n_t$ is the total number of data samples. $X=[X_1,\ldots,X_T]'$ and $Y=[Y_1,\ldots,Y_T]'$ are data samples of all the tasks, and $[\cdot]'$ stands for matrix transpose. For the dataset shift and small sample datasets based on multiple scenarios, the sparsity of variable selection via elastic net is more suitable than LASSO.

In this section, the specific models for molten iron temperature prediction are provided. Next, the product quality, the scenario states, and the upper bound of the hit rate need to be coordinated. Furthermore, product quality is a predictive variable, and the upper bound of the hit rate changes with the update of *Task t*. Therefore, both the data-driven modeling and optimization processes are dynamic in the data analytics of the process product quality. In Sec. (3.2), a quasi-convex optimization method is proposed for the dynamic data-driven modeling and optimization process.

### *3.2 Quasi-Convex Optimization Approach*

The theory of quasi-convex programming was proposed as early as the 1960s [53]. However, quasi-convex optimization did not have technical support for industrial purposes until the advent of the data analyticss. In this section, we convert the HRO problem with an upper bound to a quasi-convex optimization problem, and the proof process is briefly described in Appendix A. Then, a quasi-convex optimization algorithm is proposed to coordinate the hit rate and the model states. An initial hit rate is given based on an initial input of the scenario cluster. With clustering updating, the optimal hit rate $\theta_t^{[h]}$ of each quasi-convex optimization problem can be obtained, where $h$ represents the number of iterations. In the integrated industrial processes, the optimal hit rate corresponds to the upper limit of the hit rate required by downstream processes. In the iterative bi-level optimization process, the optimal hit rate satisfies the inequality relation in Eq. (23).

**Proposition 1:** Quasi-convex modelling via a family of convex functions (*refer to Agrawal* [54], *p*. 3.): The sublevel sets of a quasi-convex function can be represented as inequalities of convex functions. In this sense, every quasi-convex function can be represented by a family of convex functions. Considering a convex set $\mathbb{C}$, if $f:\mathbb{C}\to\mathbb{R}$ is quasi-convex, then there exists a family of convex functions $\Phi_{\theta_t^{[h]}}:\mathbb{C}\to\mathbb{R}$, indexed by $\theta_t^{[h]}\in\mathbb{R}$, such that

$$f\left(x\right)\le\theta_t^{[h]}\Leftrightarrow\phi_{\theta_t^{[h]}}\left(x\right)\le 0\Leftrightarrow f_0\left(x,\theta_t^{[h]}\right)\le 0\,. \tag{20}$$

The indicator functions for the sublevel sets of $f$,

$$\phi_{\theta_t^{[h]}}\left(x\right)=\begin{cases}0 & f_i\left(x\right)\le\theta_t^{[h]}\\ \infty & otherwise,\end{cases} \tag{21}$$

generate one such family.

According to the **Proposition 1**, we employ a family of convex functions $\phi\left(\theta_t^{[h]},X_t,Y_t;\beta\right)$ to approximate the optimal trace of the HRO problem**,** and a general quasi-convex optimization problem is defined to achieve the optimal hit rate as follows.

$$\phi_{\theta_t^{[h]}}\left(\beta;X_t,Y_t\right)\le 0, \tag{22}$$

$$st.\ \mathbf{H}(X_t,Y_t;\beta)\ge\theta_t^{[h-1]}, \tag{23}$$

$$\theta_t^{[h]}\in\mathbb{R},\ \ \dot{\theta}_l\le\theta_t^{[h]}\le\hat{\theta}_u,\ \ \phi\in\Phi\,. \tag{24}$$

In the dynamic clustering process, scenario tree reduction method [47] is used for the cluster update. The update process of scenario reduction and model reconstruction for constraint in Eq. (23) formulated as:

$$g_m^{[h+1]}:=\alpha_m^{[h]}g_m^{[h]}+\left(1-\alpha_m^{[h]}\right)g_j^{[h]}, \tag{25}$$

$$p_m^{[h+1]} := p_{\left(O_t = m^{[h+1]}\right)}, \tag{26}$$

$$|m - j| \leq 1\,. \tag{27}$$

where $g_m^{[h+1]}$ and $p_m^{[h+1]}$ are the model and observation probabilities of the reconstructed $m^{th}$ scenario for the $h$+1$^{th}$ iteration, respectively, and $\alpha_m^{[h]}$ is an optimized weighting factor of the $m^{th}$ scenario.

From the perspective of error optimization, the optimization problem (22) of the MTEN-FHMM model can be solved by convex optimization techniques, such as coordinate descent-based methods and semidefinite programming [55-57]. From the perspective of HRO, problem (22) is a quasi-convex optimization equivalently transformed from the hit rate optimization problem (8). According to **Proposition 1**, the maximization problem of the hit rate can be transformed into a family of convex subproblems to solve.

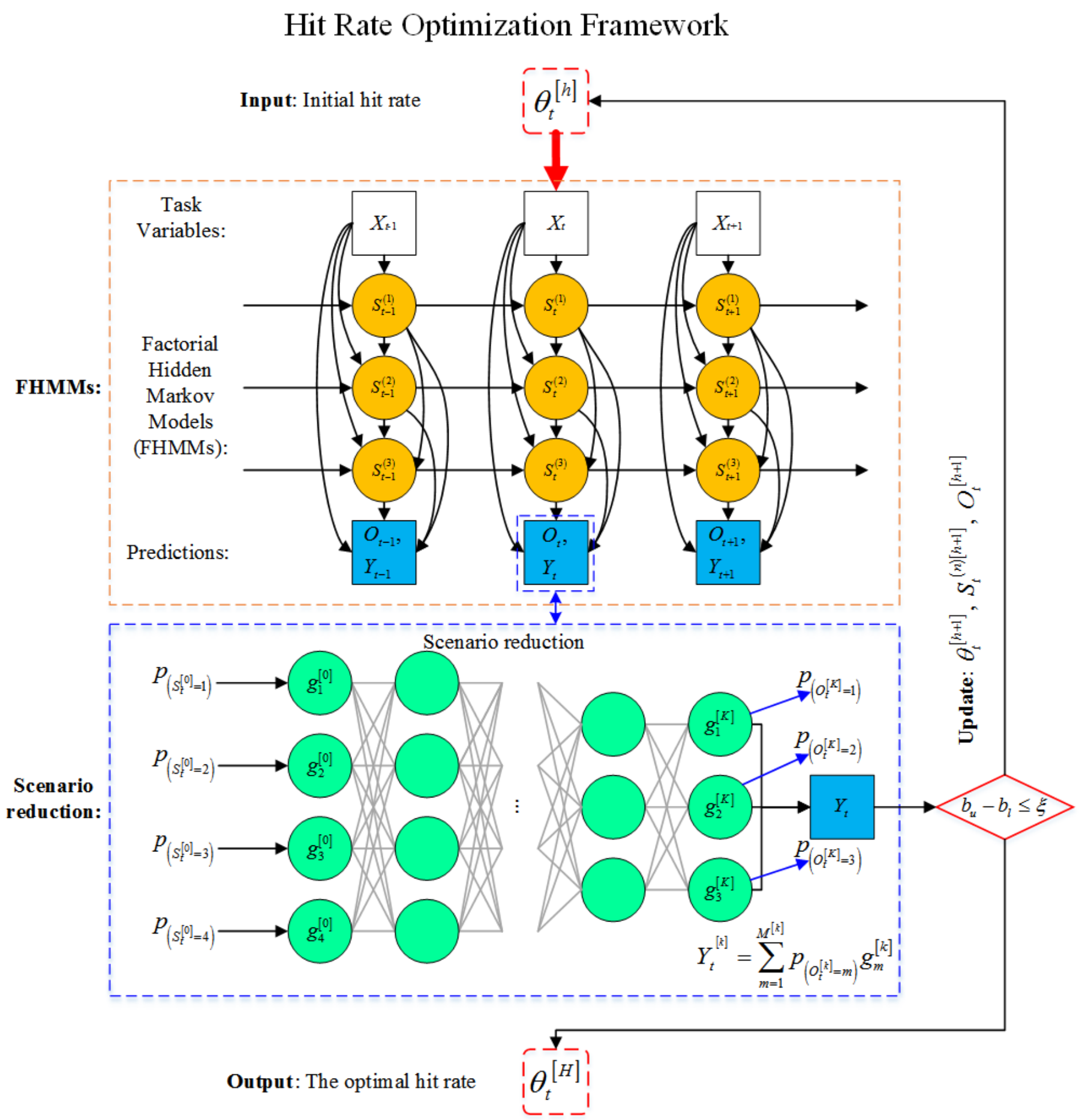

**Fig. 6.** The dynamic optimization flow of the hit rate in process product quality.

Fig. 6 illustrates the dynamic optimization process for improving the hit rate of the process product quality. **Algorithm 1** is proposed to solve the quasi-convex optimization problem of the HRO model. The quasi-convex optimization workflow of the hit rate is described as follows.

First, an initial constraint $[b_l, b_u]$, a tolerance $\xi > 0$ and an initial hit rate $\theta_t^{[h]}$ for the *Task t* are input, where $h = 0$ is an index, $h \in \mathbb{N}$.

Then, solve the convex programming problem (22).

If there is a feasible solution to the convex programming problem (22), the lower bound of the constraint on the hit rate is increased to $b_l := \theta_t^{[h]}$. Then, according to the coordinate descent method, the parameters $\alpha_m^{[h]}$ of the model reconstruction problem in Eq. (25) are solved to obtain the optimal linear combination of surrogate models and reduced clustering scenarios, which corresponds to the prediction model for $Y_t$ and model states $S_t^{(n)[h+1]}$ and $O_t^{[h+1]}$ for the FHMMs, respectively.

If optimization problem (25) is not convex feasible, the upper bound of the hit rate is reduced to $b_u := \theta_t^{[h]}$.

The above steps are repeated until the difference between the upper and lower bounds of the hit rate satisfies the stopping criteria.

Finally, the optimized hit rate $\theta_t^{[H]}$ is output.

---

**Algorithm 1** Quasi-convex optimization of the optimal hit rate modeling

---

**given** $b_l,\ b_u \in \mathbb{R}$, $0 \le b_l < b_u \le 1$, and tolerance $\xi > 0$.

**input** an initial hit rate $\theta_t^{[h]} \in [b_l, b_u]$, $h = 0,\ h \in \mathbb{N}$.

**repeat**

1. Solve the convex feasibility problem (22).
2. **if** (25) is feasible, $b_l := \theta_t^{[h]}$; **update** the model states $S_t^{(n)[h+1]}$ and $O_t^{[h+1]}$ to reconstruct the MTEN-FHMM model. **else** $b_u := \theta_t^{[h]}$.
3. $h = h + 1$, $\theta_t^{[h]} := (b_l + b_u)/2$.

**until** $b_u - b_l \le \xi$.

---

**then** $\theta_t^{[H]} = \theta_t^{[h-1]}$.

**output** the optimized hit rate $\theta_t^{[H]}$.

---

**Algorithm 1** is to explore the upper bound of the hit rate that can be achieved by the existing process when there is no sufficient knowledge. After the hit rate range of the production process is digitally twinned, the scheduling system can control the hit rate of the process product quality to a specific point or range, and the researcher can further investigate the convex feasible problem. To illustrate, since the product quality model involves solution spaces and constraints on the process variables. This leads to a further progress from a single objective optimization of hit rate to a multi-objective optimization based on control variables and conditions. The following **Algorithm 2** gives the optimization algorithm to find a frontier of the product quality function based on the controlled hit rate, where $\text{MAE}_t$ represents the mean absolute error of the time-windowed training set.

---

**Algorithm 2** Quasi-convex optimization of the specific hit rate modeling

---

**given** $MAE_t^{[0]}$**=inf** .

**input** a convex feasibility problem (22) that holds with a specific hit rate $\theta_t^{[*]}, h = 1,\ h \in \mathbb{N}$.

**repeat**

1. The grid search method is used to search the parameter $\alpha_m^{[h]}$ in Eq. (25).
2. **if** $MAE_t^{[h]} < MAE_t^{[h-1]}$, $\alpha_m^{[*]} = \alpha_m^{[h]}$; **else** $\alpha_m^{[*]} = \alpha_m^{[h-1]}$.
3. $h = h+1$.

**until** the grid search is finished.

**output** the optimized hit rate $\alpha_m^{[*]}$.

---

To summarize, the proposed quasi-convex frontier of the hit rate optimization is available when **Algorithm 2** is applied to each iteration of **Algorithm 1**. As shown in Fig. 9, the improvement in hit rate is accompanied by an increase in the model accuracy during the optimization search.

## 4. NUMERICAL EXPERIMENTS

There remains a cognitive gap between the assumption of the hidden Markov process and the nature of complex industrial processes. Therefore, the experimental section contains two numerical experiments: a quantitative experiment on the Monte Carlo simulation supported by the assumptions of the Hidden Markov process; and a numerical experiment on the prediction of the process product quality (hot metal temperature) based on the actual dataset. The Monte Carlo simulation shows the workflow of the hit rate optimization in the low-dimensional visible space and illustrates the advantages of the proposed approach. The second experiment is to analyze and validate the algorithm performance with two real BF-BOF datasets. In second experiment, we not only compare the HRO model with the baseline model, but also analyze the performance gap caused by different hypotheses on the dataset and different regularizations of the surrogate model.

### *4.1 Evaluation metrics*

In this section, the evaluation criteria include the hit rate (HR), mean absolute error (MAE), R-square ($R^2$), mean absolute percentage error (MAPE), and root-mean-square error (RMSE). HR, MAE, MAPE, $R^2$, MAPE, RMSE show the deviation between predictions and the actual value.

Combined with the hit rate optimization and the evaluation metrics, these metrics are further improved and used for the performance comparison of different models. Based on the comparison result, the simulation experiment analyzes and reveals the relations and differences between the error minimization problem and the hit rate optimization problem. Moreover, the theoretical performance and advantages of the proposed HRO method are investigated for different hit rate requirements and different data set distributions. In addition, in practical industrial applications, these metrics have the potential to be used to analyze the quasi-convex frontier, i.e., the precision upper bound for the available data quality under different assumptions, process variables and models.

The hit rate criterion in process product quality is defined by

$$HR = \frac{\sum_{i=1}^{n} \mathbf{1}\left(b_l \le y_i^p - y_i \le b_u\right)}{n} \times 100\%, \tag{28}$$

where $n$ is the number of samples in the test set, $y_i^p$ is the prediction value and $y_i$ is the measurement value. For the molten iron dispatch, the bounds of the hit range are set to ±10 °C of the HMT measured at the endpoint.

MAE [58] reflects the overall deviation between the expected value and the real output according to the predictive model for all samples. It is defined by

$$MAE = \frac{1}{n}\sum_{i=1}^{n}\left|y_i^p - y_i\right|. \tag{29}$$

RMSE [59] is a scale-dependent value and is very sensitive to large or small errors in a set of measurements. For the hot metal dispatch, RMSE is visualized to evaluate the accuracy of HMT prediction with the unit of "°C". RMSE is defined by

$$RMSE = \sqrt{\frac{1}{n}\sum_{i=1}^{n}\left(y_i^p - y_i\right)^2}. \tag{30}$$

For the integrated process industry, the process product quality within the hit range are all qualified. Therefore, the MAE and RMSE indicators are refined with the hit interval $[b_l, b_u]$, as follows.

$$MAE\left(\left[b_l, b_u\right]\right) = \frac{1}{n}\sum_{i=1}^{n}\left|\left(y_i^p - y_i\right)\mathbf{1}\left(y_i^p - y_i \geq b_u \left\| y_i^p - y_i \leq b_l\right.\right)\right|, \tag{31}$$

$$RMSE\left(\left[b_l, b_u\right]\right) = \sqrt{\frac{1}{n}\sum_{i=1}^{n}\left(\left(y_i^p - y_i\right)\mathbf{1}\left(y_i^p - y_i \geq b_u \left\| y_i^p - y_i \leq b_l\right.\right)\right)^2}. \tag{32}$$

$R^2$ [60] is a unitless parameter and can evaluate the fitting effect of the model as follows.

$$R^2 = 1 - \frac{\sum_{i=1}^{n}\left(y_i^p - \bar{y}\right)^2}{\sum_{i=1}^{n}\left(y_i - \bar{y}\right)^2}, \tag{33}$$

where $\bar{y}$ is the mean measurement value.

When the hit interval is far less than the range of the predicted objects, MAPE [61] is suitable to evaluate the limit of the method in practical engineering problems. MAPE usually presents the accuracy as a percentage. It can be calculated by

$$MAPE = \frac{1}{n}\sum_{i=1}^{n}\left(\frac{y_i^p - y_i}{y_i}\right)\times 100\%. \tag{34}$$

However, MAPE will be distorted when the true value ranges around zero. Therefore, the adjusted MAPE$_a$ is used to replace MAPE in the simulation experiment. It can be improved by

$$MAPE_a = \frac{1}{n}\sum_{i=1}^{n}\left(\frac{y_i^p - y_i}{\max\left(y_i, 1\right)}\right)\times 100\% . \quad (35)$$

### *4.2 Simulation experiments of the HMM-based multipopulation data*

This experiment is a dynamic numerical simulation experiment under two assumptions: the dynamic process of the state recognition of the model is a hidden Markov process with a single hidden Markov chain; the samples come from multiple populations, and the regression model of each population is linear. For illustration, three representative hidden Markov multipopulation experiments are elaborately prepared.

#### *4.2.1 Monte Carlo simulation setup*

According to the following steps, we designed a benchmark problem to verify and compare the effectiveness and performance of the methods.

*Step* 1: Generate a hidden Markov multipopulation dataset.

*Step* 1.1: Generate a label list of time series $\left\{S_i^{(m)} \middle| m \in \{1,2,3\}, i = 1,2,3\ldots\right\}$ based on the hidden Markov process, according to the transfer matrix $M_{trans}$ and emission matrix $M_{emis}$. Here, $i$ is a time index, and the length of the list is 1000.

*Step* 1.2: Generate data samples with the label list.

$$\left\{\left(\left(x_i, y_i\right), S_i^{(m)}\right)\right\},\ \ i = 1,2,3,\ldots, \quad (36)$$

where $i$ is the time index, $x_i$ represents the observable variable, and the variable $y_i$ and label $S_i$ cannot be observed before time $i$. The labels $\{S_i\}$ divide data samples $\left\{\left(x_i, y_i\right)\right\}$ into three categories, corresponding to three datasets $\left\{\left(x_i^{(m)}, y_i^{(m)}\right) \middle| \left(x_i\right)^{(m)} \sim N\left(\mu^{(m)}, \sigma^{(m)}\right), m \in \{1,2,3\}\right\}$. Fig. 7 shows the true distributions of the data samples of the three sets mapped on the $x$-$y$ plane.

*Step* 2: Set the width of the hit range.

$$I(x_i, y_i) = \begin{cases} 1, & \left| x_i^T \hat{\beta}_i - y_i \right| \le e \\ 0, & else \end{cases} \quad (37)$$

*Step* 3: Modelling. The first 500 samples of the hidden Markov multipopulation sequence are used as the training set to establish a linear regression model with the ordinary least squares (OLS) method.

*Step* 4: Test. The next 500 samples are used as the test set to be predicted.

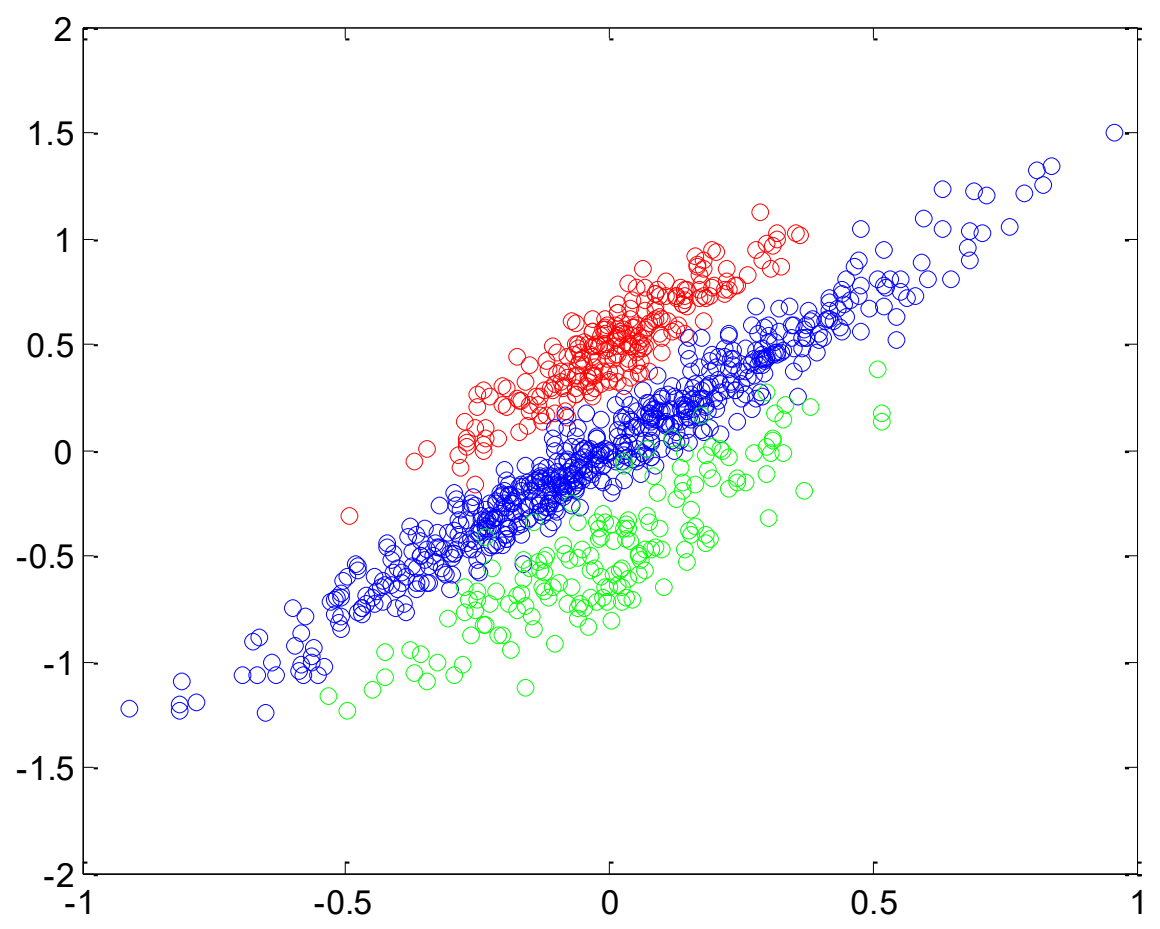

**Fig. 7.** The *x*-*y* projection of functional label clustering.

*4.2.2 Comparison experiment*

For comparison, a baseline experiment and two controlled experiments are presented. Parameter settings for the baseline group are shown in Table 1. The parameter design refers to the process of steel production. For example, there are multiple hidden data populations with different noises via the BF-BOF route. The reason for this phenomenon lies in the different equipment conditions and operation modes of different production lines. Two controlled experiments are designed by changing the intervals between the centroids and the width of the hit range. The detailed simulation settings are as follows

*a) Baseline group:*

**Table 1.** Parameter setting of the baseline experiment based on hidden Markov process

| Parameters | Value |
|---|---|
| Distribution of the populations | $(x_i)^{(1)} \sim N\left(0, 0.15^2\right)$<br>$(x_i)^{(2)} \sim N\left(0, 0.3^2\right)$<br>$(x_i)^{(3)} \sim N\left(0, 0.2^2\right)$ |
| Noise | $noise_1 = N\left(0, 0.1^2\right)$<br>$noise_2 = N\left(0, 0.2^2\right)$ |
| Regression model | $Y_1 = 1.5X_1 + 0.5 + noise_1$<br>$Y_2 = 1.5X_2 + noise_1$<br>$Y_3 = 1.5X_3 - 0.5 + noise_2$ |
| Transfer matrix | $M_{\text{trans}} = \begin{bmatrix} 0.4 & 0.55 & 0.05 \\ 0.15 & 0.7 & 0.15 \\ 0.05 & 0.55 & 0.4 \end{bmatrix}$ |
| Emission matrix | $M_{\text{emis}} = \begin{bmatrix} 0.55 & 0.4 & 0.05 \\ 0.2 & 0.7 & 0.1 \\ 0.05 & 0.4 & 0.55 \end{bmatrix}$ |
| Width of the hit range | $I\left(x_i, y_i\right) = \begin{cases} 1, & \left\lvert x_i^T \hat{\beta}_i - y_i \right\rvert \le 0.2 \\ 0, & else \end{cases}$ |
| $card\left(\left\{\left(x_i, y_i\right)_1\right\}\right)$ | 242 |
| $card\left(\left\{\left(x_i, y_i\right)_2\right\}\right)$ | 602 |
| $card\left(\left\{\left(x_i, y_i\right)_3\right\}\right)$ | 156 |

*b) Controlled group 1:*

Let the intervals among the three populations be smaller (*i.e.* $y_1 = 1.5x_1 + 0.3 + noise$ and $y_3 = 1.5x_3 - 0.3 + noise$ ), and the other parameters remain the same as those in the baseline experiment.

*c) Controlled group 2:*

Let the width of the hit range be smaller (*i.e.* $I(x_i, y_i) = \begin{cases} 1, & \left| x_i^T \hat{\beta}_i - y_i \right| \le 0.1 \\ 0, & else \end{cases}$), and the other parameters remain the same as in controlled group 1.

*4.2.3 Results and analysis*

In the comparison experiment, two multipopulation learning problems are involved: the HMM learning problem and the HRO problem. The result of HMM is optimized by the Baum-Welch algorithm. Without the dynamic feature extraction and sparse optimization, the proposed HRO model is simplified by combining the HMM model and HRO framework (HMM-HRO). The result of the HRO is calculated by the maximum hit rate.

Fig. 8 illustrates the prediction results of the baseline group and two controlled groups. From Table 2, the six evaluation criteria are calculated to analyze the performance between the HMM and HMM-HRO models. Although there is a gap between HMM classification and the real situation, it shows that the number and centroids of populations can be correctly identified with HMM in Fig. 8(*a*-2). For the baseline setup, the quasi-convex property of the baseline problem can be seen. At this point, the degree of scenario reconstruction is low, as shown in Fig. 8(*a*-3); Table 2 shows that there is no significant difference between the optimal hit rate of the HMM model and HMM-HRO model, and the $MAE_{hr}$ can be further optimized by the HMM-HRO model.

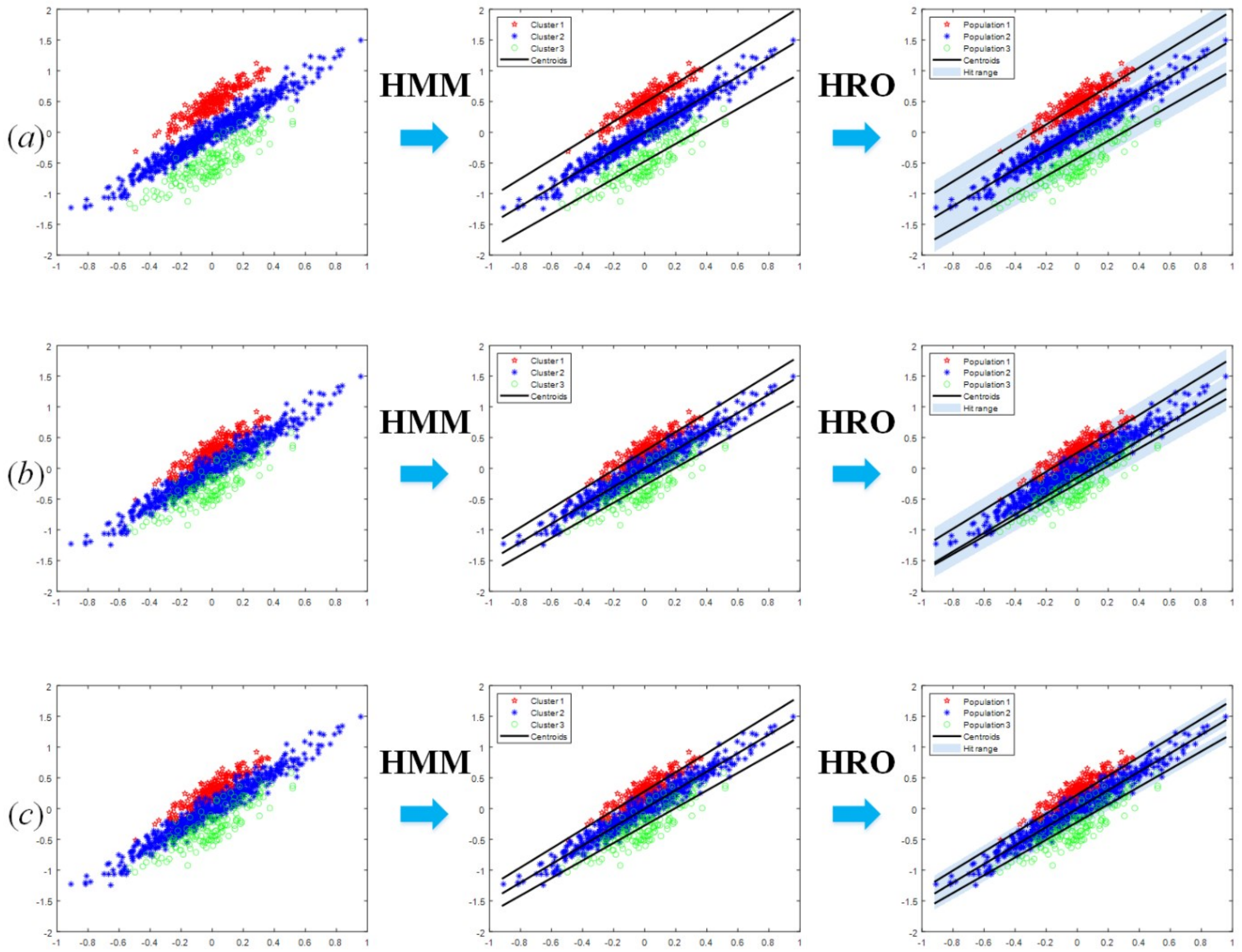

**Fig. 8.** Results comparison: (*a*) baseline group, (*b*) controlled group 1 and (*c*) controlled group 2.

**Table 2.** Performance between HMM and HRO

| Dataset | Criteria | HMM_train | HRO_train | HMM_test | HRO_test |
|---|---|---|---|---|---|
| Baseline group | $\mathrm{HR}([-0.2, 0.2])$ | 73.00% | 73.60% | 71.60% | 71.60% |
| | MAE | 0.1696 | 0.1691 | 0.1748 | 0.1738 |
| | RMSE | 0.2459 | 0.2442 | 0.2516 | 0.2496 |
| | $R^2$ | 0.7749 | 0.7779 | 0.7471 | 0.7511 |
| | $\mathrm{MAPE}_a$ | 16.92% | 16.88% | 17.45% | 17.36% |
| | $\mathrm{MAE}([-0.2, 0.2])$ | 0.1150 | 0.1128 | 0.1216 | 0.1205 |
| | $\mathrm{RMSE}([-0.2, 0.2])$ | 0.2332 | 0.2309 | 0.2395 | 0.2373 |
| Controlled group 1 | $\mathrm{HR}([-0.2, 0.2])$ | 78.60% | 83.80% | 75.80% | 82.60% |
| | MAE | 0.1280 | 0.1165 | 0.1327 | 0.1155 |
| | RMSE | 0.1702 | 0.1453 | 0.1749 | 0.1447 |
| | $R^2$ | 0.8603 | 0.8982 | 0.8375 | 0.8887 |
| | $\mathrm{MAPE}_a$ | 12.77% | 11.63% | 13.25% | 11.53% |
| | $\mathrm{MAE}([-0.2, 0.2])$ | 0.0657 | 0.0434 | 0.0732 | 0.0457 |
| | $\mathrm{RMSE}([-0.2, 0.2])$ | 0.1476 | 0.1101 | 0.1539 | 0.1118 |
| Controlled group 2 | $\mathrm{HR}([-0.1, 0.1])$ | 52.80% | 54.60% | 49.40% | 49.80% |
| | MAE | 0.1280 | 0.1301 | 0.1327 | 0.1360 |
| | RMSE | 0.1702 | 0.1742 | 0.1749 | 0.1811 |
| | $R^2$ | 0.8603 | 0.8537 | 0.8375 | 0.8257 |
| | $\mathrm{MAPE}_a$ | 12.77% | 12.98% | 13.25% | 13.58% |
| | $\mathrm{MAE}([-0.1, 0.1])$ | 0.1027 | 0.1040 | 0.1107 | 0.1143 |
| | $\mathrm{RMSE}([-0.1, 0.1])$ | 0.1653 | 0.1692 | 0.1709 | 0.1774 |
| | $\mathrm{HR}([-0.2, 0.2])$ | 78.60% | 76.60% | 75.80% | 74.20% |
| | $\mathrm{MAE}([-0.2, 0.2])$ | 0.0657 | 0.0721 | 0.0732 | 0.0796 |
| | $\mathrm{RMSE}([-0.2, 0.2])$ | 0.1476 | 0.1543 | 0.1539 | 0.1623 |

In order to explain the proposed quasi-convex frontier, the controlled group 1 deserves attention. In order to balance the prediction accuracy, the cost function and the constraints of the process variables, the hit rate in practical operations need to be controlled in a middle and feasible point/range, as shown in **Table 3.** And then, the trajectory of hit rate with the mean absolute error optimization in the quasi-convex optimization process forms a quasi-convex frontier of the hit rate in Fig. 9.

**Table 3.** The results of the controlled hit rate

| Model | Criteria | Controlled HR=0.7 | Controlled HR=0.725 | Controlled HR=0.75 | Controlled HR=0.775 | Controlled HR=0.8 |
|---|---|---|---|---|---|---|
| Control experiment I (train) | HR | 70.20% | 72.40% | 75.20% | 77.00% | 80.00% |
| | MAE | 0.1490 | 0.1421 | 0.1359 | 0.1299 | 0.1244 |
| | RMSE | 0.1918 | 0.1817 | 0.1737 | 0.1652 | 0.1544 |
| | $R^2$ | 0.8225 | 0.8408 | 0.8545 | 0.8683 | 0.8850 |
| | $MAPE_a$ | 14.87% | 14.18% | 13.56% | 12.97% | 12.41% |
| | $MAE_{hr}$ | 0.0918 | 0.0820 | 0.0726 | 0.0652 | 0.0539 |
| | $RMSE_{hr}$ | 0.1735 | 0.1609 | 0.1503 | 0.1398 | 0.1231 |
| Control experiment I (test) | HR | 70.60% | 73.00% | 75.00% | 77.20% | 80.00% |
| | MAE | 0.1499 | 0.1427 | 0.1377 | 0.1316 | 0.1248 |
| | RMSE | 0.1948 | 0.1843 | 0.1771 | 0.1681 | 0.1548 |
| | $R^2$ | 0.7985 | 0.8197 | 0.8334 | 0.8499 | 0.8726 |
| | $MAPE_a$ | 14.97% | 14.25% | 13.75% | 13.14% | 12.45% |
| | $MAE_{hr}$ | 0.0921 | 0.0819 | 0.0743 | 0.0658 | 0.0543 |
| | $RMSE_{hr}$ | 0.1760 | 0.1630 | 0.1539 | 0.1423 | 0.1239 |

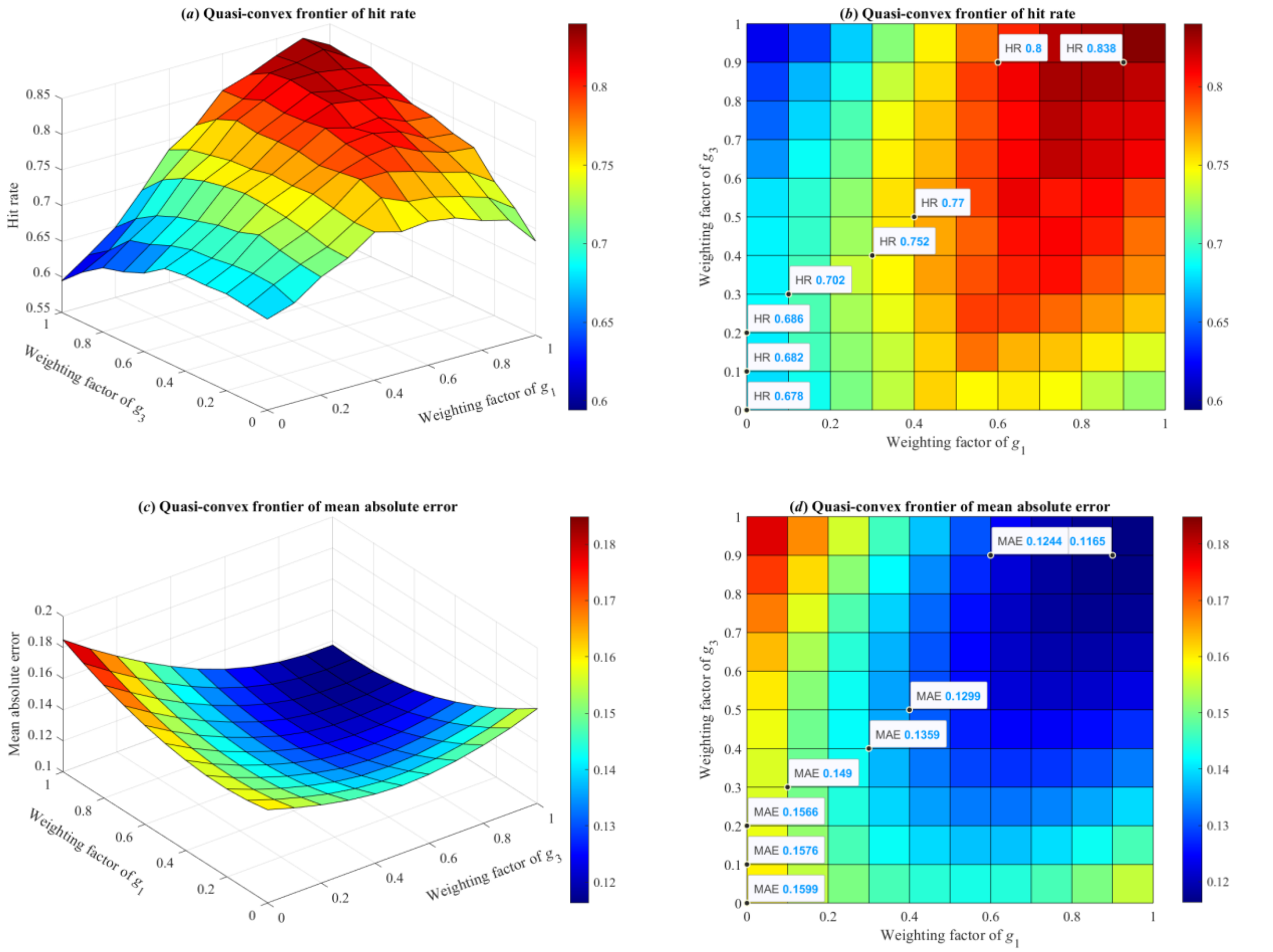

**Fig. 9.** Quasi-convex frontier of hit rate optimization: *a*) quasi-convex frontier of hit rate; *b*) optimization trajectory (Quasi-convex frontier) of hit rate; *c*) quasi-convex frontier of corresponding mean absolute error; *d*) optimization trajectory (Quasi-convex frontier) of mean absolute error. ($g_1$, $g_2$ and $g_3$ represent the surrogate models of process product quality under three scenarios, respectively)

In Fig. 9, it can be seen that the optimization trends in the iterative optimization process are roughly similar for both the hit rate metric and the MAE metric. The surfaces in Figs. 8(*a*) and (*c*) are the optimal hit rate profiles based on the weighting factors of the different scenario models. In this article, the profile data are named as quasi-convex frontier. Depending on the molten iron scheduling operation experience, it can be found that the quasi-convex frontier contains a Quasi-convex frontier composed by a number of marked points, as shown in Figs. 8(*b*) and (*d*). The points on the Quasi-convex frontier separately correspond to an operation optimization strategy based on the optimal hit rate and the corresponding control variables. The points below the Quasi-convex frontier correspond to some feasible and controllable hit rate operation strategies. The points beyond the Quasi-convex frontier correspond to infeasible operational strategies, and the corresponding model parameters are considered to be deteriorated. The deterioration of model optimization is common in high-dimensional spatiotemporal statistical modeling. This statistical phenomenon is further explained and discussed in Sec. (5).

In order to visualize the quasi-convex frontier and the Quasi-convex frontier, the simulation experiments were developed. While actual industrial high-dimensional data modeling projects have numbers of hidden scenarios, complex variable relationships, and inevitable cognitive gaps existing in intermediate processes and statistical assumptions (Markov process assumption). Therefore, the quasi-convex frontier and Quasi-convex frontier are not smooth enough to visualize the process of quasi-convex optimization in low-dimensional space.

Compared with the baseline experiment, the interval between the populations decreased in the controlled group 1. When the hit rate requirement remained unchanged, some scenario functions gradually became closer. As a result, the scenario functions in Fig. 8(*b*-3) are shifted, and two of the scenarios are nearly merged compared to Fig. 8(*a*-3). Under this condition, the HR, MAE and $R^2$ of the HMM+HRO method are significantly higher than those of the HMM method. Controlled group 2 decreased the hit interval based on the condition of controlled group 1, the merged scene functions are separated again in Fig. 8(*c*-3). Compared with the other two experiments, the hit rate dropped significantly, but the difference of the other evaluation indicators between the two methods was not obvious.

In addition, the distribution of the scenarios approximates the true distribution of the populations. Fig. 10 shows the error density distribution based on single-population function estimation. For the baseline group, there are three populations in the baseline experiment, and the populations are separable. In this case, the subpopulations can be estimated by Bayesian variational inference [62-63,31]. For the controlled group 1 and 2, the optimization method is suitable for the subpopulation model to achieve the optimal evaluation criteria [64].

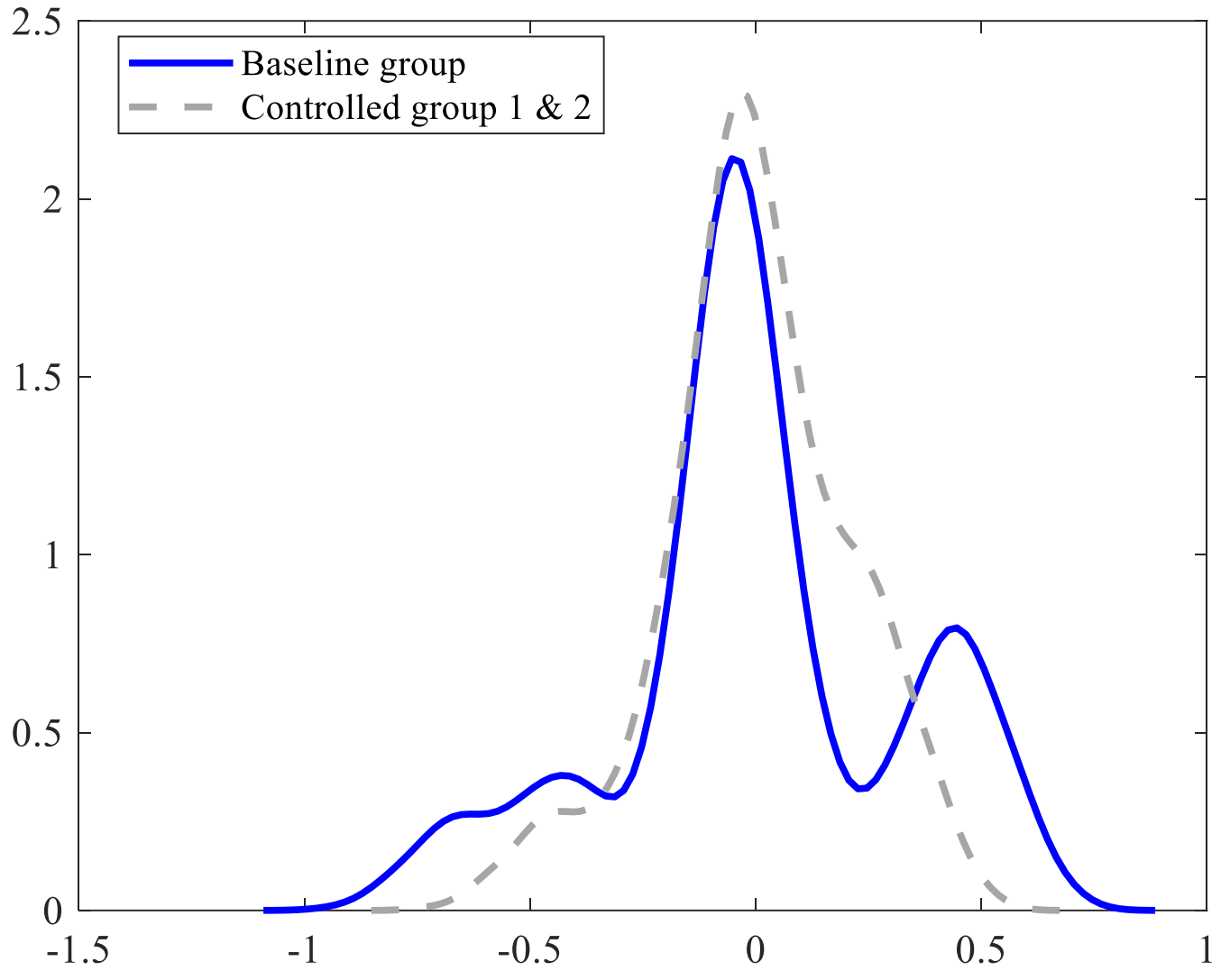

**Fig. 10.** Density distribution function of single-population multi-linear regression.

### *4.3 Hit rate optimization of hot metal temperature prediction via the BF-BOF route*

#### *4.3.1 Experimental setup*

Experiment II involved the actual tasks of the BF-BOF route in a Chinese steel plant in 2020. Dataset A and Dataset B collect 1,000 sequential tasks from two different BF-BOF routes. The maximum number of hidden scenarios is set to three. In the initial iteration, the historical datasets A and B were divided into three scenarios based on the error intervals of the multiple linear regression model with the OLS optimization method (*i.e.,* (-∞, -10), [-10, 10], (10, +∞)). The model training is carried out in a rolling way. In each iteration of training, the samples of the most recent 500 BF-BOF production tasks are used as the training set, and the samples of next 50 tasks are used as the test set.

The input of the product quality model includes continuous variables and categorical variables. When constructing elastic net model, the categorical variables need to be converted to dummy variables. In the hot metal temperature prediction problem, the input features of all the experimental models involved 16 process variables and 1 constant term, including 2 categorical variables.

#### *4.3.2 Comparison experiment*

In the comparison experiment, the research object is the HRO model and its baseline model is the FHMM model (baseline model). The hypotheses for the datasets are divided into single-population and multipopulation. The employed single-population models for the comparison experiment include the OLS model with the feature space of quadratic polynomial variables, a two-layer autoencoder (AE-2) model [65] with 32 neurons per layer and a least squares support vector machine (LS-SVM) model [66]. The OLS model (classical model) is also a baseline model representing the classical effective regression model with the minimum error optimization in practical applications. Compared with the baseline model (FHMM), the results highlight the proposed the research significance of the multipopulation dataset shift problem in industrial engineering problems.

In Sec. (4.2), it is discussed that in addition to the stochastic process, the main reason that affects the prediction accuracy is the gap between the surrogate model and the real model. In Sec. (4.3), the comparison between the multitask LASSO (MTLASSO) and MTEN models are used to illustrate the impact of the regularization in the surrogate modeling.

*4.3.3 Results and analysis*

For the rolling training dataset, the number of input variables and the corresponding proportions for the three scenario models are shown in Fig. 11. It can be seen that both the sparsity and variable selection of the proposed model change dynamically with the dataset shifting. The corresponding percentages dynamically change as well. The dynamic variable selection with adaptive regularization for the surrogate modeling process are detailed in Appendix B.

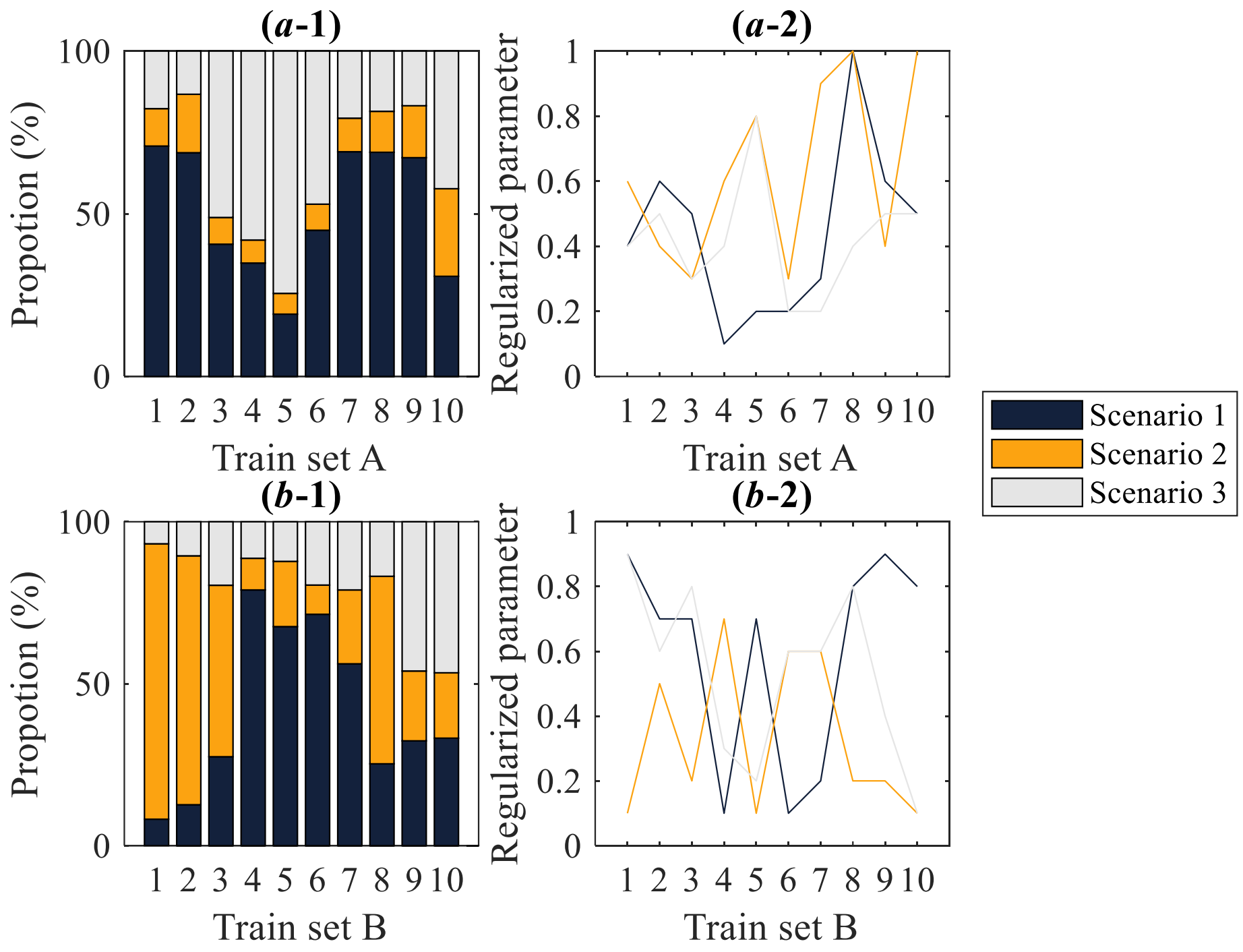

**Fig. 11.** Input variable proportions and parameter tuning of the three scenario models: *a*-1) proportion of variables selected for training set A with moving time window; *a*-2) regularization parameters of the elastic network models with time window movement of training set A; *b*-1) proportion of variables selected for training set B with moving time window; *b*) regularization parameters of the elastic network models with time window movement of training set B.

The performance comparison of the prediction accuracy of the six models is performed and shown in Table 4. The evaluation criteria verify the rationality of the proposed HRO method for optimizing the hit rate of the HMT prediction via BF-BOF. For better illustration, the prediction of the first 50 test samples is shown in detail in Figs. 12-15(a), and the complete results are shown in Figs. 12-15(b). Fig. 16 records the hit rate curve for the tasks in dataset A and B.

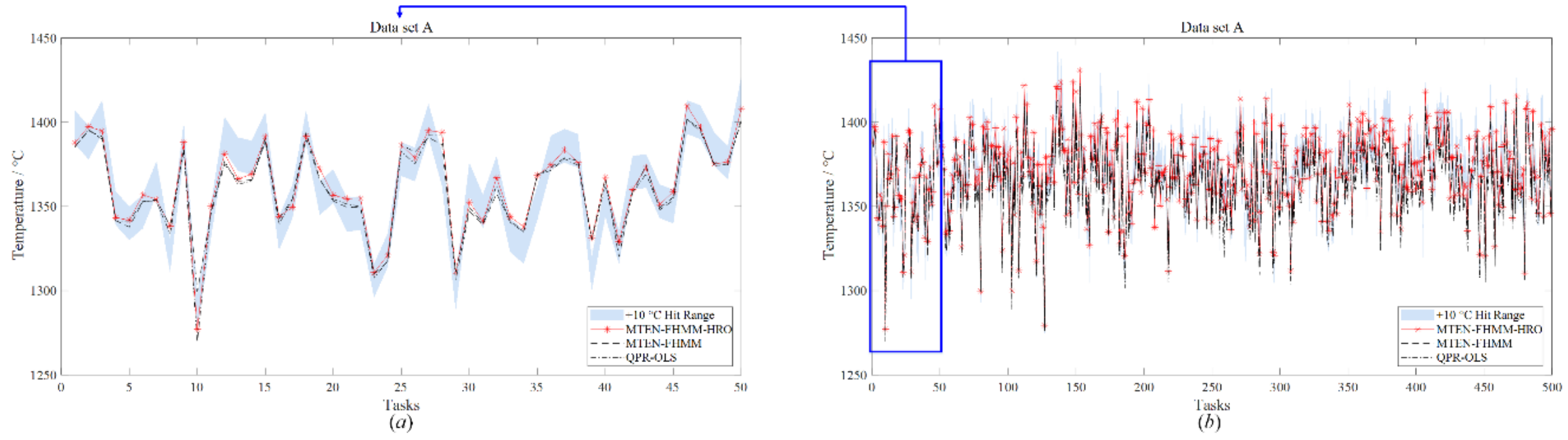

**Fig. 12.** Results comparison of test set A with the HRO model, FHMM model and OLS model.

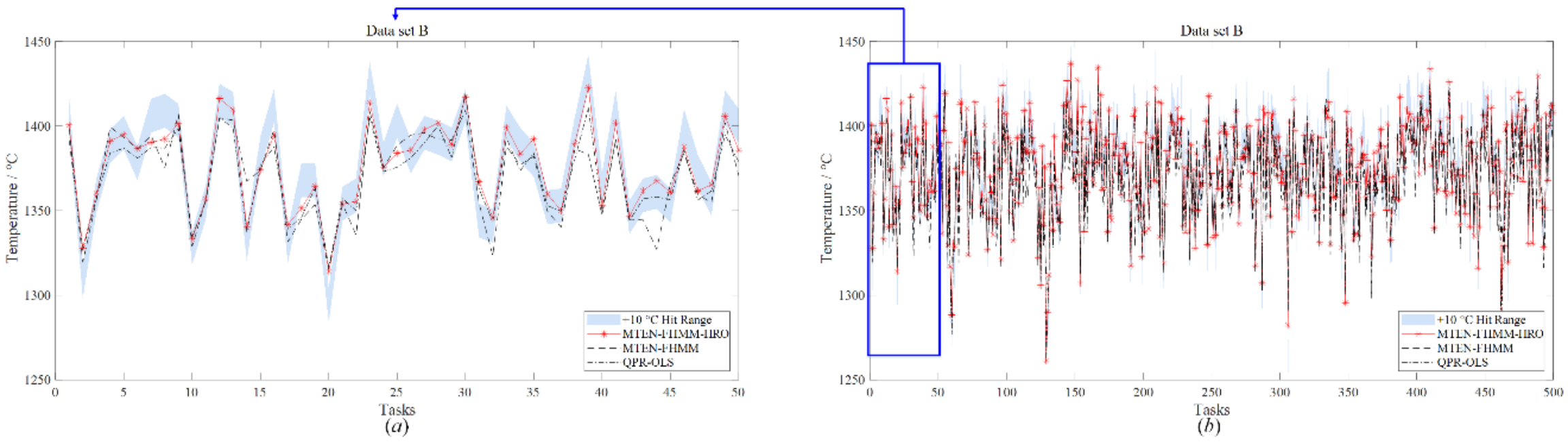

**Fig. 13.** Results comparison of test set B with the HRO model, FHMM model and OLS model.

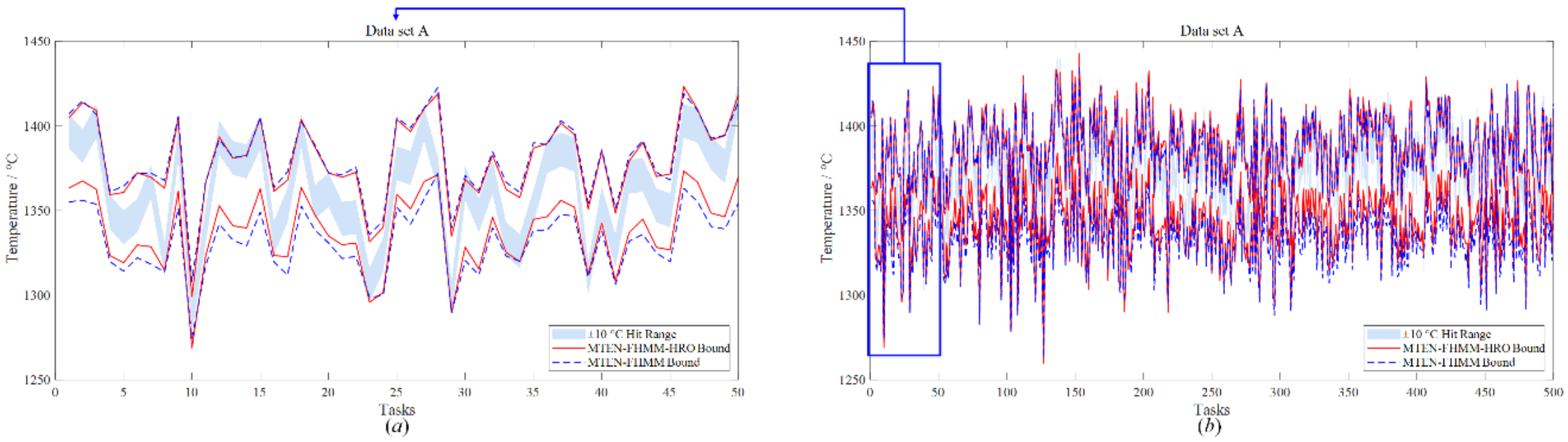

**Fig. 14.** Upper bound and lower bound comparison between the HRO model and FHMM model of test set A.

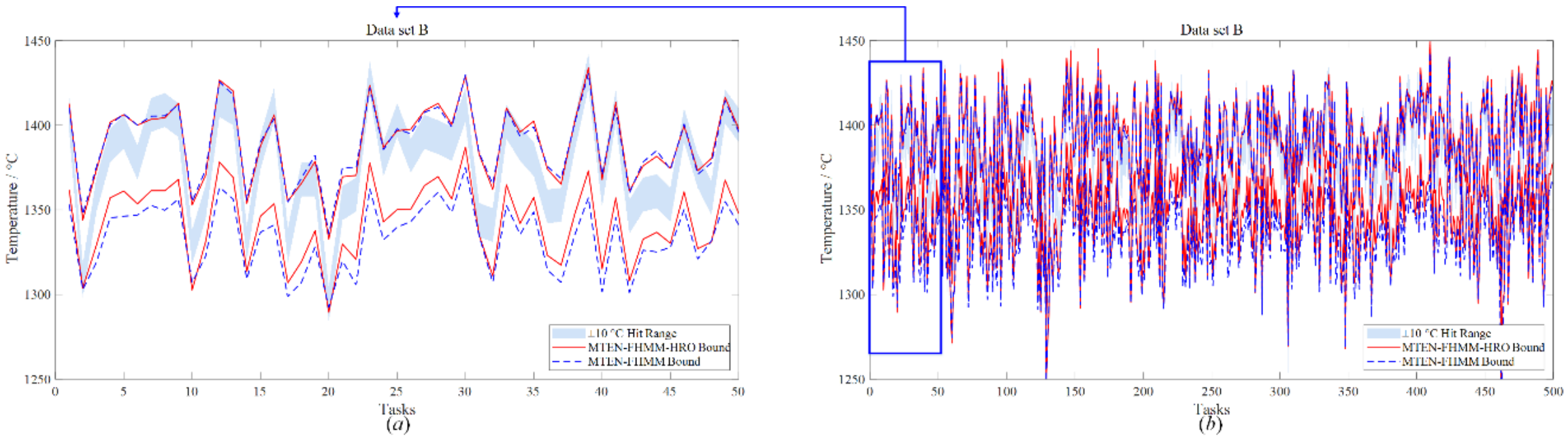

**Fig. 15.** Upper bound and lower bound comparison between the HRO model and FHMM model of test set B.

**Table 4.** Performance of Different Models for End-point HMT Prediction

| Dataset | Hypothesis | Model | ±10 ℃ Hit Rate | $MAE_{hr}$ | $RMSE_{hr}$ | R2 | MAPE |
|---|---|---|---|---|---|---|---|
| Test set A | Multi-population | HRO (MTEN) | 75.60% | 3.7974 | 7.9550 | 0.8641 | 0.55% |
| | | FHMM (MTEN) | 73.40% | 3.9185 | 7.8325 | 0.8724 | 0.53% |
| | | FHMM (MTLASSO) | 65.80% | 5.4934 | 9.7425 | 0.8223 | 0.63% |
| | Single-population | OLS | 53.20% | 8.1574 | 12.5921 | 0.7315 | 0.78% |
| | | LS-SVM | 51.00% | 8.2875 | 12.4145 | 0.7423 | 0.76% |
| | | AE-2 | 26.20% | 21.8812 | 27.9955 | -0.2175 | 1.69% |
| Test set B | Multi-population | HRO (MTEN) | 75.20% | 3.9806 | 8.1757 | 0.9094 | 0.56% |
| | | FHMM (MTEN) | 70.60% | 4.1991 | 7.9615 | 0.9195 | 0.52% |
| | | FHMM (MTLASSO) | 63.00% | 5.9258 | 10.1043 | 0.8786 | 0.66% |
| | Single-population | OLS | 57.40% | 6.9877 | 11.2168 | 0.8593 | 0.70% |
| | | LS-SVM | 45.00% | 11.2188 | 17.1267 | 0.6955 | 0.97% |
| | | AE-2 | 18.20% | 28.9625 | 35.5026 | -0.2504 | 2.17% |

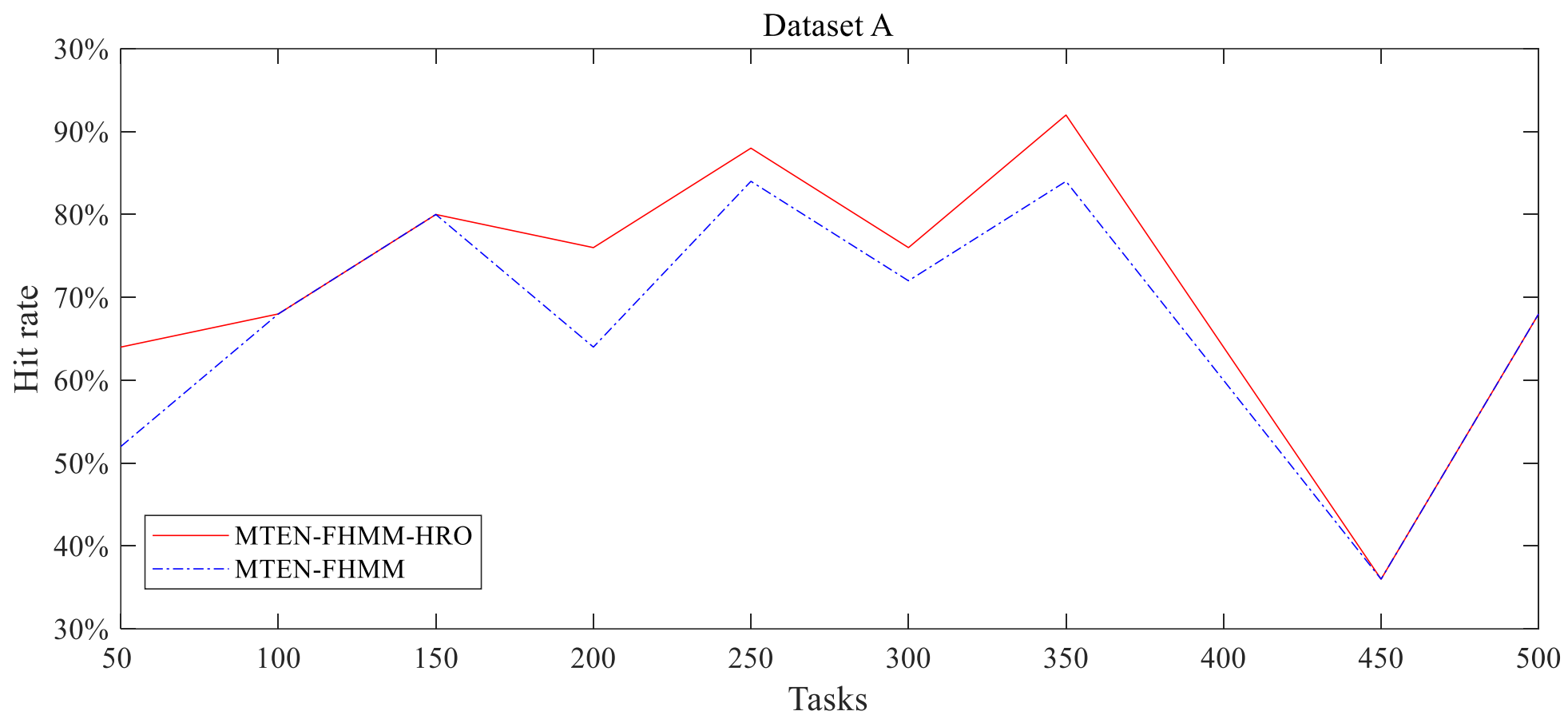

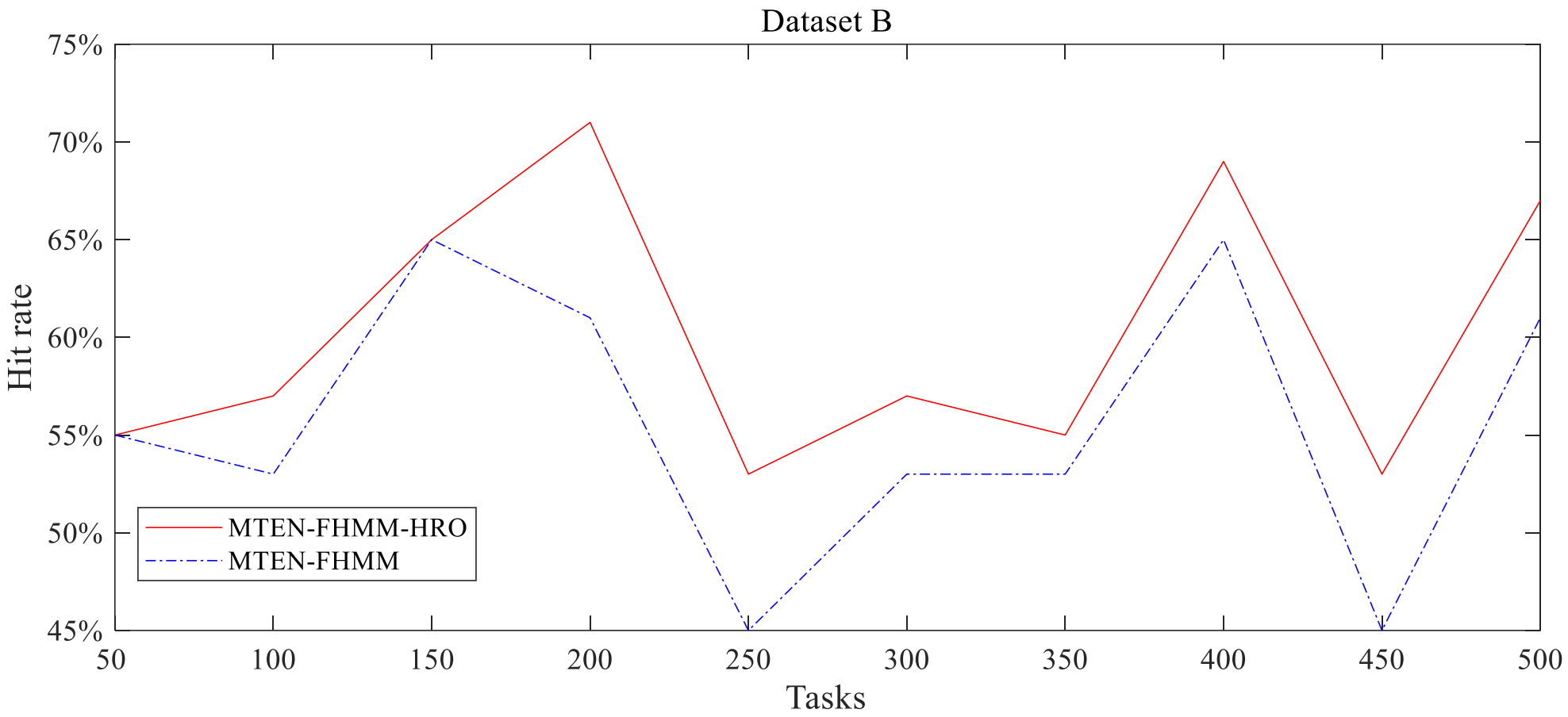

**Fig. 16.** Hit rate curve of test set A & B.

From the perspective of the optimal hit rate, the ±10 ℃ hit interval of the HRO model via the two BF-BOF routes both reach the optimal value compared with other models. In particular, the hit rate of the HRO model is improved by at least 3.00% and 6.52%, respectively, compared to the proposed baseline models, and by at least 42.11% and 31.01%, respectively, compared to the classical models in practical applications. The proposed data-driven quasi-convex method for hit rate optimization balances the relationship among the deviation of the surrogate model, the sample distribution and the noise of different data populations, and the characteristic of the stochastic process to reach the optimal hit rate in Figs. 11-15. Figs. 11 and 12 show that the difficulty of prediction with the hit rate requirement has risen from the convex problem to the nonconvex problem, where precision is required from a traditional point to a qualified range. After the FHMM model is optimized by the HRO, it can be seen that the top bound and the bottom bound of the surrogate models shrink to the hit range in Figs. 14 and 15. In Fig. 16 the curve of the hit rate verifies that the proposed method can guarantee stability for the HMT prediction problem. And occasionally the error optimization also achieves the optimal hit rate.

From the perspective of error optimization, the proposed HRO model shows the best performance on $MAE_{hr}$ in datasets A and B, while it follows the HRO model on the other error indicators. In terms of $RMSE_{hr}$, since the hit range (±10°C) is much smaller than the general temperature of molten iron, the result depends on the sensitivity of the RMSE function. Furthermore, the $R^2$ values of the HRO and FHMM models reach 0.8641/0.9094 and 0.8724/0.9195, respectively, which verifies the fitting effect of the proposed multipopulation modelling. The MAPE represents the gap between the data-driven surrogate models and the ideal model. Compared with the gap of the MTLASSO model, the MTEN model decreased the gap from 0.63%-0.53% and 0.66%-0.52%, respectively. Such a result is commonly observed when the number of features is greater than the sample size or there are many conflicting data instances in the training set. Based on the experimental results, it can be concluded that the multiple population datasets are more appropriate to describe the HMT via the black-box BF-BOF route by comparing performance indicators in Table 3. Among the single-population models, AE-2 has not exploited its powerful generalization capability. As a representative multilayer neural network model, the results of AE-2 also represent that most of existing neural network models suffer from model generalization ability after deprived of sufficient high-quality data support, thus leading to the deterioration of their optimization results. On the other hand, the results also demonstrate the complexity and difficulty of real datasets. Oriented to the real data streams in industry, the OLS method is proven to be the optimal and conservative choice for most of the practical industrial applications available today.

In summary, the experimental results show that the proposed data-driven quasi-convex method is more accurate and robust for the hit rate optimization of the process product quality in integrated steel production. Furthermore, the proposed method is a quasi-convex optimization method that can achieve the optimal hit rate and provide the quasi-convex frontier that corresponds to the Quasi-convex frontier of multi-objective optimization.

## 5. DISCUSSION

In statistics, error propagation for high-dimensional spatiotemporal data regression [67,68] and generalization capabilities of the surrogate models for operational optimization [69-71] are popular research directions on data interpretability. It is being extensively noticed in artificial intelligence and smart manufacturing domains, such as deep networks and reinforcement learning. However, restricted by the complexity and visualization of high-dimensional spatiotemporal problems, these research directions have not been fully investigated and cognized. The multi-stage black-box industrial process with surrogate models studied in this article presents a typical example of that statistical phenomenon in industry, which elaborates 1) the cognitive gap between three layers of nature, model and data, respectively; 2) the error propagation process across the surrogate models of process product quality; and 3) the reasons for the substandard degree of whole-process optimization and terminal product quality based on individual process optimizations in operational optimization. The specific procedure is schematically shown below.

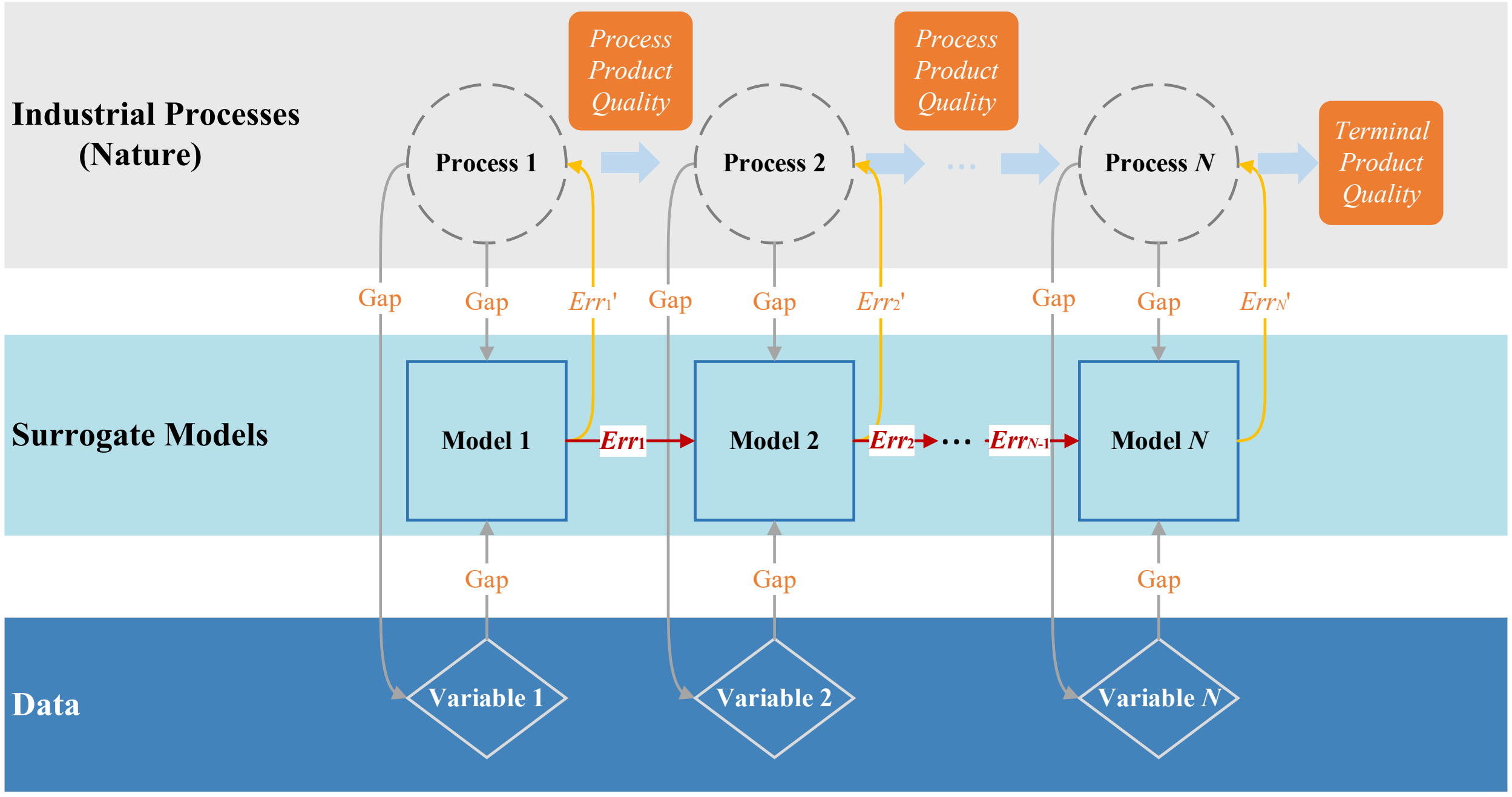

**Fig. 17.** Schematic diagram of the impact of error propagation and cognitive gaps on product quality.

There are three layers of control and optimization for the process industry in Fig. 17: industrial processes, surrogate models, and data. The error propagation chain $Err_1 \to Err_2 \to \cdots \to Err_N$ is the key to the industrial process integration and the improvement of product quality since the scheduling plan of the downstream process is performed based on the predictions of the upstream model. Another error propagation chain $Err'_1 \to Err'_2 \to \cdots \to Err'_N$ influences the process product quality of the individual processes. In addition, it can be seen that there is an inevitable gap in human cognition or modeling techniques between the three layers. And, the gap is being gradually approached with the thorough research on macro to micro mechanisms.

The experimental chapter presents specific visual illustrations and comparative results for the above complex problem of hit rate optimization based on both simulated and real datasets. The error propagation of optimization for a single-stage error can be seen in Fig. 9. The solution with the minimum error is probably the deteriorated position in the hit rate optimization profile. Therefore, along with the error propagation, the final output of the multi-stage process is a scheduling scheme with the amplified errors and deteriorated operational variables for the hit rate and the product quality. As a result, the hit rate of the process product quality and the yield of the final product are not controllable and predictable. In Fig. 17, the relationship between the hit rate and the prediction error for the real dataset. In comparison, by controlling the error propagation chain within the feasible hit interval between two adjacent processes, the hit rate optimization method is significantly more beneficial to maintain the accuracy, stability and safety of the process product quality in the whole process.

## 6. CONCLUSION

In this article, a data-driven quasi-convex approach for the hit rate optimization is proposed for the process product quality prediction with the hit rate requirement in integrated steel production. The hit rate optimization is an important optimization problem with application value but is not easy to solve. The nonconvexity in mathematical optimization and the difficulty in black-box modeling are discussed specifically. In order to find an effective solution, the data-driven models of the product quality are established by a collection of surrogated models. Then, all the surrogated models are unified into the proposed dynamic optimization framework. As the model state probability is estimated, the surrogated models are reconstructed to achieve the optimal hit rate. Finally, the theoretical and practical performance of the proposed approach is verified by simulation experiments and two actual steel production datasets, respectively. On the actual datasets, the proposed method improves the hit rate indicator by at least 42.11% and 31.01%, respectively, compared with the classical models, and by at least 3.00% and 6.52%, respectively, compared with the proposed baseline models.

The future research focuses on the robust control of data outside the hit interval. The cognitive gaps between the true nature and the mechanism research, and the gaps between the real industrial processes and mathematical tools (such as differential manifolds and deep learning with constraints) will be explored theoretically with the upgrading of hardware and software of industry.

**APPENDIX A**

An indicator function is defined as follows,

$$\mathbf{I}\left(f\left(x\right)\right)=\begin{cases}0, & f\left(x\right)\in\mathbb{C};\\ 1, & f\left(x\right)\notin\mathbb{C}.\end{cases} \tag{A-1}$$

where $f\left(x\right):\mathbb{C}\rightarrow\mathbb{R}$ is a convex function and $\mathbb{C}$ is a convex set. Then, the indicator function is proven to be a convex function (*refer to Boyd* [72], *p*. 68, *example* 3.1).

The hit rate function proposed in this paper is shown as follows.

$$\mathbf{I}\left(f\left(\beta\right)\right)=\begin{cases}0, & f\left(\beta\right)\notin\left[b_l,b_u\right];\\ 1, & f\left(\beta\right)\in\left[b_l,b_u\right].\end{cases} \tag{A-2}$$

where $\left[b_l,b_u\right]$ is a convex set. When $\mathbf{I}\left(f\left(\beta\right)\right)=0$, $f\left(\beta\right)$ belongs to a concave set.

By definition of the convex function, the indicator function $\mathbf{I}\left(f\left(\beta\right)\right)$ is proven to be a concave function. Then, a composite function of the indicator function $\mathbf{I}_{\mathbb{C}}\left(\beta\right)$ is also a concave function, shown as follows.

$$\mathbf{I}_{\mathbb{C}}\left(\beta\right)=\frac{\sum\mathbf{I}\left(f\left(\beta\right)\right)}{n} \tag{A-3}$$

Then, the maximizing the objective function corresponds to a minimizing problem as follows.

$$\underset{\beta}{min}\quad -\mathbf{I}_{\mathbb{C}}\left(\beta\right) \tag{A-4}$$

where $-\mathbf{I}_{\mathbb{C}}\left(\beta\right)$ is the convex function.

Then, the objective function of the HRO problem is defined as

$$\mathbf{I}_{\mathbb{C}}\left(x\right)=\frac{\sum\mathbf{I}\left(f\left(x\right)\right)}{n}\leq\hat{\theta} \tag{A-5}$$

By **Definition 1** of the quasi-convex function, the objective function (6-5) is a quasi-concave function for the maximizing problem (8) and it corresponds to a quasi-convex function for the minimizing problem (6-4).

**Definition 1** (See Boyd [72], p. 95, definition): A function $f:\mathbb{R}\rightarrow\mathbb{R}$ is called *quasi-convex (or unimodal)* if its domain and all its sublevel sets

$$S_\alpha=\left\{x\in\mathbf{dom}\,f\,\middle|\,f\left(x\right)\leq\alpha\right\},$$

for $\alpha \in \mathbb{R}$ , are convex. A function is *quasi-concave* if $-f$ is quasi-convex, *i.e.*, every superlevel set $\{x \mid f(x) \geq \alpha\}$ is convex. A function that is both quasi-convex and quasi-concave is called quasi-linear. If a function $f$ is quasi-linear, then its domain, and every level set $\{x \mid f(x) = \alpha\}$ is convex.

This article introduced the FHMM-MTEN model as the surrogate model, which satisfies the above assumption of the convex function $f(x)$. In addition, according to the requirement of hit rate in actual industrial production, the hit rate or the process yield rate can also be controlled within a certain range. Similar to the optimization process for maximizing hit rate, it requires changing the upper and lower bound of hit rate constraints in **Algorithm 1**.

## APPENDIX B

**Table B-1.** Variable Selection for Dataset A

| T | SCE1_VAR_NUM | $\lambda$ | SCE2_VAR_NUM | $\lambda$ | SCE3_VAR_NUM | $\lambda$ |
|---|---|---|---|---|---|---|
| 1 | 80 | 0.4 | 13 | 0.6 | 20 | 0.4 |
| 2 | 88 | 0.6 | 23 | 0.4 | 17 | 0.5 |
| 3 | 89 | 0.5 | 18 | 0.3 | 112 | 0.3 |
| 4 | 69 | 0.1 | 14 | 0.6 | 115 | 0.4 |
| 5 | 27 | 0.2 | 9 | 0.8 | 105 | 0.8 |
| 6 | 84 | 0.2 | 15 | 0.3 | 88 | 0.2 |
| 7 | 87 | 0.3 | 13 | 0.9 | 26 | 0.2 |
| 8 | 93 | 1 | 17 | 1 | 25 | 0.4 |
| 9 | 80 | 0.6 | 19 | 0.4 | 20 | 0.5 |
| 10 | 24 | 0.5 | 21 | 1 | 33 | 0.5 |

**Table B-2.** Variable Selection for Dataset B

| T | SCE1_VAR_NUM | $\lambda$ | SCE2_VAR_NUM | $\lambda$ | SCE3_VAR_NUM | $\lambda$ |
|---|---|---|---|---|---|---|
| 1 | 12 | 0.9 | 125 | 0.1 | 10 | 0.9 |
| 2 | 12 | 0.7 | 73 | 0.5 | 10 | 0.6 |
| 3 | 14 | 0.7 | 27 | 0.2 | 10 | 0.8 |
| 4 | 105 | 0.1 | 13 | 0.7 | 15 | 0.3 |
| 5 | 94 | 0.7 | 28 | 0.1 | 17 | 0.2 |
| 6 | 95 | 0.1 | 12 | 0.6 | 26 | 0.6 |
| 7 | 32 | 0.2 | 13 | 0.6 | 12 | 0.6 |
| 8 | 24 | 0.8 | 55 | 0.2 | 16 | 0.8 |
| 9 | 33 | 0.9 | 22 | 0.2 | 47 | 0.4 |
| 10 | 74 | 0.8 | 45 | 0.1 | 104 | 0.1 |

* When the number of selected variables is less than the minimum value given by the mechanism analysis (the minimum number in this experiment is 17), the surrogate model is recognized as underfitting. For surrogate modeling in this case, the more stable regression model is employed in practice as a conservative strategy, and the underfitting model is substituted by either a linear regression model or a polynomial regression model.
* The variable selection was inspired by the work of Ouyang [73] and Wang [19].